\documentclass[preprint,12pt,authoryear]{elsarticle}

\usepackage{setspace}
\usepackage{nicefrac}
\usepackage{graphicx}
\graphicspath{{images/}}
\usepackage[colorlinks]{hyperref}
\usepackage{multirow}
\usepackage{amsmath}
\usepackage{amssymb}
\usepackage{mathtools}
\usepackage{float}
\usepackage{hhline}
\usepackage{boldline}
\usepackage{adjustbox}
\usepackage{ctable}
\usepackage{caption}
\usepackage{hyperref}
\usepackage{ctable}
\usepackage{placeins}
\usepackage{xspace}
\usepackage{lineno}

\hypersetup{bookmarks=false,         
}

\newcommand{\segnet}{\textsc{segnet}\xspace}
\newcommand{\segneth}{\textsc{seg$\cdot$h}\xspace}
\newcommand{\hed}{\textsc{hed}\xspace}
\newcommand{\hedh}{\textsc{hed$\cdot$h}\xspace}
\newcommand{\fcn}{\textsc{fcn}\xspace}
\newcommand{\fcns}{\textsc{fcn}s\xspace}
\newcommand{\vgg}{\textsc{vgg}\xspace}
\newcommand{\dlreth}{\textsc{fcn$\cdot$h}\xspace}
\newcommand{\dlrethVGG}{\textsc{fcn$\cdot$h-v}\xspace}
\newcommand{\dlrethPAS}{\textsc{fcn$\cdot$h-p}\xspace}
\newcommand{\deeplab}{\textsc{deeplab}\xspace}

\journal{of }

\begin{document}

\begin{frontmatter}

\title{Classification with an edge: improving semantic image
  segmentation with boundary detection}

\author{D. Marmanis\textsuperscript{a,c},
 	 K. Schindler\textsuperscript{b},
  	 J. D. Wegner\textsuperscript{b},
	 S. Galliani\textsuperscript{b},
 	 M. Datcu\textsuperscript{a},
 	 U. Stilla\textsuperscript{c}
}

\address{\textsuperscript{a }DLR-IMF Department, German Aerospace Center, Oberpfaffenhofen, Germany --						  	\{dimitrios.marmanis, mihai.datcu \}@dlr.de\\

	\textsuperscript{b }Photogrammetry and Remote Sensing, ETH Zurich, Switzerland --	\{konrad.schindler, jan.wegner, silvano.galliani \}@geod.baug.ethz.ch\\

	\textsuperscript{c }Photogrammetry and Remote Sensing, TU M\"unchen, Germany -- stilla@tum.de
}

\begin{abstract}

We present an end-to-end trainable deep convolutional neural network
(DCNN) for semantic segmentation with built-in awareness of
semantically meaningful boundaries.
Semantic segmentation is a fundamental remote sensing task, and most
state-of-the-art methods rely on DCNNs as their workhorse. A major
reason for their success is that deep networks learn to accumulate
contextual information over very large receptive fields.
However, this success comes at a cost, since the associated loss of
effective spatial resolution washes out high-frequency details and
leads to blurry object boundaries.
Here, we propose to counter this effect by combining semantic
segmentation with semantically informed edge detection, thus making
class boundaries explicit in the model.
First, we construct a comparatively simple, memory-efficient model by
adding boundary detection to the \segnet encoder-decoder
architecture. Second, we also include boundary detection in
\textsc{fcn}-type models and set up a high-end classifier ensemble.
We show that boundary detection significantly improves semantic
segmentation with CNNs in an end-to-end training scheme.  Our best
model achieves $>90$\% overall accuracy on the ISPRS Vaihingen
benchmark.
\end{abstract}

\end{frontmatter}

\section{Introduction}\label{introduction}

Semantic image segmentation (a.k.a.\ landcover classification) is the
process of turning an input image into a raster map, by assigning every
pixel to an object class from a predefined class nomenclature.
Automatic semantic segmentation has been a fundamental problem of
remote sensing data analysis for many
years~\citep{Fu1969,Richards2013}.
In recent years there has been a growing interest to perform semantic
segmentation also in urban areas, using conventional aerial images or
even image data recorded from low-flying drones.
Images at such high resolution (GSD 5-30$\,$cm) have quite different
properties.
Intricate spatial details emerge like for instance road markings, roof
tiles or individual branches of trees, which increase the spectral
variability within an object class. On the other hand, the spectral
resolution of sensors is limited to three or four broad bands
so spectral material signatures are less distinctive.
Hence, a large portion of the semantic information is encoded in the
image texture rather than the individual pixel intensities, and much
effort has gone into extracting features from the raw images that make
the class information
explicit~\citep[e.g.][]{Franklin1993,Barnsley1996,DallaMura2010,Tokarczyk2015}.

At present the state-of-the-art tool for semantic image segmentation,
in remote sensing as well as other fields of image analysis, are deep
convolutional neural networks (DCNNs).%
\footnote{For example, the 15 best-performing participants in the
  \emph{ISPRS Vaihingen} benchmark, not counting multiple entries from
  the same group, all use DCNNs.} %
For semantic segmentation one uses so-called fully convolutional
networks (\textsc{fcn}s), which output the class likelihoods for an entire
image at once. \textsc{fcn}s have become a standard tool that is readily
available in neural network software.

\begin{figure}[t!]
\centering

\label{fig:teaser}
\begin{tabular}{lll}
	  \includegraphics[scale=0.3]{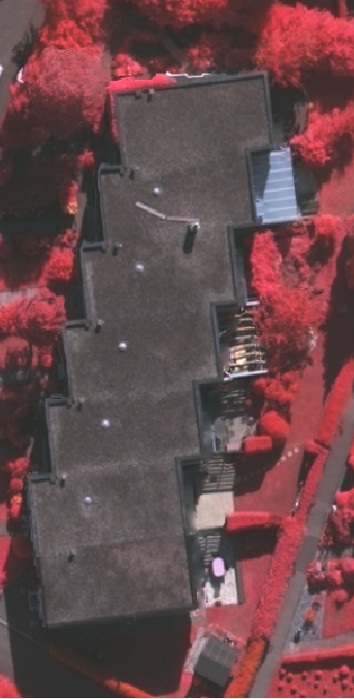}
 	& \includegraphics[scale=0.3]{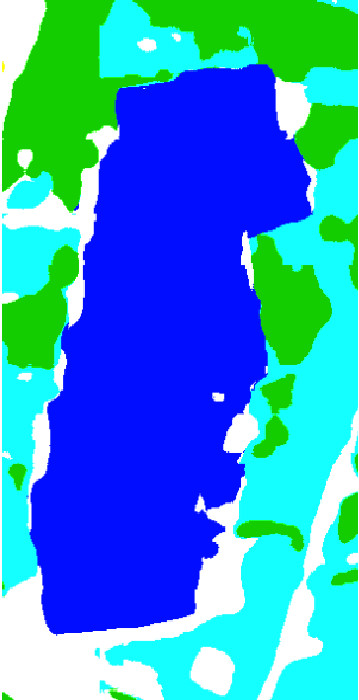}
	& \includegraphics[scale=0.3]{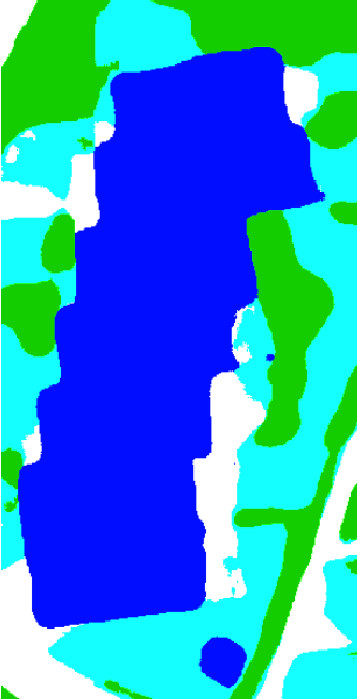}
\end{tabular}
\caption{Semantic segmentation. Left: input image. Middle: DCNN
    segmentation, object boundaries tend to be blurred. Right: We
    propose to mitigate this effect by including an explicit object
    boundary detector in the network.}
\end{figure}

Why are DCNNs so successful (if given sufficient training data and
computing resources)? Much has been said about their ability to learn
the complete mapping from raw images to class labels (``end-to-end
learning''), thus making heuristic feature design obsolete.
Another strength is maybe even more important for their excellent
performance: deep networks capture a lot of \emph{context} in a
tractable manner. Each convolution layer combines information
from nearby pixels, and each pooling layer enlarges the footprint
of subsequent convolutions in the input image. Together, this means
that the output at a given pixel is influenced by a large spatial
neighborhood.
When the task is pixel-wise semantic segmentation%
\footnote{As opposed to, e.g., object recognition or speech
  recognition.}, %
their unparalleled ability to represent context however comes at a
price. There is a trade-off between strong downsampling, which allows
the network to see a large context, but loses high-frequency detail;
and accurate localization of the object boundaries, which requires
just that local detail.
We note that in generic computer vision with close-range images that
effect is much less critical. In a typical photo, say a portrait or a
street scene, there are few, big individual objects and only few
object boundaries, whose precise location moreover is only defined up
to a few pixels.
In some cases segmentation is even defined as finding the
boundaries between very few (say, $<5$) dominant
regions~\citep{russell2009segmenting} that make up the image.
On the contrary, for our remote sensing task we expect at least tens to
hundreds of small segments, such as individual cars, trees, etc.

There has been some research that tries to mitigate the blurring of
boundaries due to down-sampling and subsequent up-sampling, either by
using the a-trous convolution (dilated convolution)
~\citep{Yu2016multiscale,ChenPK0Y16,sherrah2016fully}
or by adding skip connections from early to deep layers of the
network, so as to reintroduce the high-frequency detail after
upsampling~\citep{Dosovitskij2015, badrinarayanan2017segnet,
marmanis2016semantic}.
Still, we find that when applied to remote sensing data with many small
objects, \fcns tend to blur object
boundaries and visually degrade the result (see Fig.1).

In this paper, we propose to explicitly represent \emph{class-boundaries}
in the form of pixel-wise contour likelihoods, and to
include them in the segmentation process. By class-boundaries we mean
the boundaries between regions that have different semantic class,
i.e., we aim for a ``semantic edge-detection''.
Our hypothesis is that if those boundaries, which by definition
correspond to the location of the label transitions, are made
available to the network, then it should learn to align the
segmentation to them.

Importantly, recent work, in particular the \emph{holistically nested
  edge detection} \citep[\hed][]{xie2015holistically,
  Kokkinos2016pushing} has shown that edge detection can also be
formulated as a \textsc{fcn}, reaching excellent results.
We can therefore merge the two tasks into a single network, train them
together and exploit synergies between them.
The result is an end-to-end trainable model for semantic segmentation
with a built-in awareness of semantically meaningful boundaries.
We show experimentally that explicitly taking into account class
boundaries significantly improves labeling accuracy, for our datasets
up to $6$\%.

Overall, our boundary-aware ensemble segmentation network reaches
state-of-the-art accuracy on the ISPRS semantic labeling
benchmark.
In particular, we find that adding boundary detection consistently
improves the segmentation of man-made object classes with well-defined
boundaries. On the contrary, we do not observe an improvement for
vegetation classes, which have intrinsically fuzzy boundaries at our
target resolution.
Moreover, our experiments suggest that integrated boundary detection
is beneficial both for light encoder/decoder architectures with
comparatively few parameters like \segnet, and for high-performance
networks with \emph{fully connected layers}, such as the \textsc{vgg}
family, which are much heavier to train in terms of memory usage and
computation time.

Our tests also confirm, perhaps unsurprisingly, that DCNNs perform optimally when
merged into \textit{ensemble models}. Combining multiple semantic
segmentation networks seems beneficial to reduce the bias of
individual models, both when using the same architecture with
different initializations, and when using different model
architectures with identical initializations.

In terms of practical relevance, a main message of this paper is that,
with DCNNs, semantic segmentation is practically usable also for very
high resolution urban remote sensing.
It is typically thought that object extraction algorithms are good
enough for (possibly semi-automatic) applications when their
predictions are at least 80\% correct~\citep{mayer2006}.
In our experiments, the $F_1$-score (harmonic mean between precision
and recall) surpasses this threshold for all object classes, including
small ones like cars. Frequent, well-defined man-made classes reach
well above 90\%.

\section{Related Work}\label{sec:related}

Semantic segmentation has been a core topic of computer vision as well
as remote sensing for many years. A full review is beyond the scope of
this paper, we refer the reader to textbooks such
as~\citep{Szeliski2010,Richards2013}. Here, we concentrate on methods
that employ neural networks.

Even in the early phase of the CNN revival, semantic segmentation
was tackled by \citet{grangier2009deep}. The study investigates
progressively deeper CNNs for predicting pixel labels from a local
neighborhood, and already shows very promising results, albeit at
coarse spatial resolution.
\citet{SocherEtAl2012:CRNN} start their semantic segmentation pipeline
with a single convolution and pooling layer. On top of that ``feature
extractor'' they stack recursive neural networks (RNN), which are
employed on local blocks of the previous layer in a rather convolutional
manner. The RNNs have random weights, there is no end-to-end
training. Notably, that work uses RGB-D images as input and processes
depth in a separate stream, as we do in the present work.
Related to this is the approach of \citet{pinheiro2014recurrent}, who
use ``recurrent CNNs'', meaning that they stack multiple shallow
networks with tied convolution weights on top of each other, at each
level using the predicted label maps from the previous level and an
appropriately down-scaled version of the raw image as input.
\citet{farabet2013learning} train CNNs with $3$ convolutional layers and
a fully connected layer for semantic segmentation, then post-process
the results with a CRF or by averaging over super-pixels to obtain a
smooth segmentation. Like in our work, that paper generates an image
pyramid and processes each scale separately with the CNN, but in
contrast to our work the filter weights are tied across scales.
An important milestone was the \emph{fully convolutional network}
(\textsc{fcn}) of \citet{long2015fully}. In that work it was shown
that the final, fully connected layers of the network can be seen as a
large stack of convolutions, which makes it possible to compute
spatially explicit label maps efficiently.
An further important work in this context is the \emph{Holistically-
Nested Edge Detection} (\textsc{hed}) of~\citet{xie2015holistically}, who
showed that an \fcn\ trained to output edge maps instead of class
labels is also an excellent edge detector. Their network was
initialized with the \vgg object detection network, so arguably the
edge detection is supported by the semantic information captured in
that network.
Variants of \hed have been explored by other authors,
\citep{Kokkinos2016pushing} confirming that CNNs are at present also
the state of the art for edge detection.

To undo the loss of spatial resolution due to the pooling layers of
\textsc{fcn}, \citet{noh2015learning} propose to add an unpooling and
upsampling (``deconvolution'') network~\citep{zeiler2014visualizing}
on top of it. The result is a sort of encoder-decoder structure that
upsamples the segmentation map back to the resolution of the input
image.
\citet{yang2016object} employ a similar strategy for the opposite
purpose: their primary goal is not semantic labeling but rather a
``semantically informed'' edge detection which accentuates edges that
lie on object contours.
Also related is the work of \citet{bertasius2015high}. They find
candidate edge pixels with conventional edge detection, read out the
activations at those pixels from the convolution layers of the
(frozen) \vgg object detection network, and separately train a
classifier on the vector of activations to separate object boundaries
from other edges.
For semantic segmentation, it has also been proposed to additionally
add skip connections from lower layers in the encoder part (before
pooling) to corresponding decoder layers (after unpooling) so as to
re-inject high-frequency image contours into the upsampling process
~\citep{Dosovitskij2015, marmanis2016semantic}.
This architecture has been simplified by the \segnet model of
\citet{badrinarayanan2017segnet}: the fully connected layers are
discarded, which drastically reduces the number of free
parameters. Moreover, that architecture makes it possible to keep
track of pixel indices during max-pooling and restore the values to
the correct position during unpooling.

In the context of individual object detection it has been
proposed to train the encoder/detector part first, freeze it, and
train the decoder part separately~\citep{pinheiro2016learning}.
The \deeplab network of \citet{chen2015semantic-a} explores a different
upsampling strategy: low-resolution output from the \fcn\ is first
upsampled bilinearly, then refined with a fully connected
CRF~\citep{Kraehenbuehl2011} whose pairwise potentials are modulated
by colour differences in the original image.
Later \deeplab was extended to simultaneously learn edge
detection and semantic segmentation~\citep{chen2015semantic}. This is
perhaps the work most closely related to ours, motivated by the same
intuition that there are synergies between the two tasks, because
object boundaries often coincide with edges.
Going even further, \citet{dai2016instance} construct a joint network
for detecting object instances, assigning them to a semantic class,
and extracting a mask for each object -- i.e., per-object class
boundaries. The method is very efficient for well-localized compact
objects (``things''), since the object instances can be detected first
so as to restrict subsequent processing to those regions of
interest. On the other hand, it appears less applicable to remote
sensing, where the scene is to a large extent composed of objects
without a well-defined bounding box (``stuff'').

Regarding applications in remote sensing, shallow neural networks were
already used for semantic segmentation before the advent of deep
learning, e.g.~\citet{Bischof1992} use a classical multi-layer
perceptron to predict the semantic class of a pixel from a small
neighborhood window.
Shallow architectures are still in use:
\citet{malmgren2015convolutional} train a relatively shallow CNN with
3 convolution layer and 2 pooling layers to classify pixels in SAR
images.
\citet{langkvist2016classification} make per-pixel predictions with
shallow CNNs (where the convolution weights are found by clustering
rather than end-to-end training) and smooth the results by averaging
over independently generated super-pixels.
The mentioned works predict individually for each pixel, on the
contrary~\citet{mnih2010learning} have designed a shallow, fully
connected network for patch-wise prediction of road pixels.
An also relatively shallow \fcn\ with 3 convolution layers and 1
pooling layer is used in~\citep{saito2016multiple}.

In the last few years, different deep CNN variants have been proposed
for semantic segmentation of remote sensing images.
\citet{paisitkriangkrai2015effective} learn three separate 6-layer
CNNs that predict semantic labels for a single pixel from three
different neighborhoods. The scores are averaged with those of a
conventional random forest classifier trained on per-pixel features,
and smoothed with a conditional random field.
\citet{marcu2016dual} design a network for patchwise 2-class
prediction. It takes as input patches of two different sizes (to
represent local and global context), passes them through separate deep
convolutional architectures, and combines the results in three deep,
fully connected layers to directly output 16$\times$16 patches of
pairwise labels.

More often, recent works adopt the \fcn architecture. Overall, the
results indicate that the empirical findings from computer vision
largely translate to remote sensing images. Both our own work
\citep{marmanis2016semantic} and \citet{sherrah2016fully} advocate a
two-stream architecture that learns separate convolution layers for
the spectral information and the DSM channel, and recommend to start
from pretrained networks for the spectral channels. Our work further
supports the practice of training multiple copies of the same CNN
architecture and averaging their results~\citep{marmanis2016semantic},
and~\citet{sherrah2016fully} reports that the a-trous convolution
trick slightly mitigates the information loss due to pooling, at the
cost of much larger (40$\times$) computation times.
\citet{mou2016spatiotemporal} prefer to avoid upsampling altogether,
and instead combine the coarser semantic segmentation with a
super-pixel segmentation of the input image to restore accurate
segmentation boundaries (but not small objects below the scale of the
\fcn output). Of course, such a strategy cannot be trained end-to-end
and heavily depends on the success of the low-level super-pixel
segmentation.

A formal comparison between per-pixel CNNs and \fcns has been carried
out by \citet{volpi2017dense}. It shows advantages for \textsc{fcn}, but
unfortunately both networks do not attain the state of the art,
presumably because their encoder-decoder network lacks skip
connections to support the upsampling steps, and has been trained from
scratch, losing the benefit of large-scale pretraining.
A similar comparison is reported in \citep{kampffmeyer2016semantic},
with a single upsampling layer, and also trained from scratch. Again
the results stay below the state-of-the-art but favor
\textsc{fcn}. Median-balancing of class frequencies is also tested, but seems to
introduce a bias towards small classes. An interesting aspect of that
paper is the quantification of the network's prediction uncertainty,
based on the interpretation of drop-out as approximate Bayesian
inference~\citep{gal2016dropout}. As expected, the uncertainty is highest
near class-contours.

\section{The Model}\label{sec:method}

In the following, we describe our network architecture for
boundary-aware semantic segmentation in detail.
Following our initial hypothesis, we include edge detection early in
the process to support the subsequent semantic labeling.
As further guiding principles, we stick to the deep learning paradigm
and aim for models that can be learned end-to-end; we build on
network designs, whose performance has been independently confirmed;
and, where possible, we favor efficient, lean networks with
comparatively few tunable weights as primary building blocks.

When designing image analysis pipelines there is invariably a
trade-off between performance and usability, and DCNNs are no
exception. One can get a respectable and useful result with a rather
elegant and clean design, or push for maximum performance on the
specific task, at the cost of a (often considerably) more complex and
unwieldy model.
In this work we explore both directions: on the one hand, our basic
model is comparatively simple with $8.8\cdot10^7$ free parameters
(\hedh{}+\segneth, single-scale; see description below), and can be
trained with modest amounts of training data.
On the other hand, we also explore the maximum performance our
approach can achieve. Indeed, our high-end model, with multi-scale
processing and ensemble learning, achieves $>$90\% overall accuracy on
the ISPRS Vaihingen benchmark. But diminishing returns mean that this
requires a more convoluted architecture with $9\times$ higher memory
footprint and training time.

Since remote sensing images are too large to pass through a CNN, all
described networks operate on tiles of $256\times 256$
pixels. We classify overlapping tiles with
three different strides (150, 200, and 220 pixels) and sum the
results.

We start by introducing the building blocks of our model, and then
describe their integration and the associated technical details.
Throughout, we refrain from repeating formal mathematical definitions
for their own sake. Equation-level details can be found in the
original publications. Moreover, we make our networks publicly
available to ensure repeatability.%
\footnote{\scriptsize{\url{https://github.com/deep-unlearn/ISPRS-Classification-With-an-Edge}}}


\subsection{Building blocks}

\subsubsection*{\segneth encoder-decoder network}

\segnet \citep{badrinarayanan2017segnet} is a crossbreed between a
fully convolutional network~\citep{long2015fully} and an
encoder-decoder architecture. The input image is first passed through
a sequence of convolutions, \emph{ReLU} and $max$-pooling layers.
During $max$-pooling, the network tracks the spatial location of the
winning maximum value at every output pixel.  The output of this
\emph{encoding} stage is a representation with reduced spatial
resolution. That ``bottleneck'' forms the input to the \emph{decoding}
stage, which has the same layers as the encoder, but in reverse
order. Max-pooling layers are replaced by \emph{unpooling}, where the
values are restored back to their original location, then convolution
layers interpolate the higher-resolution image.

Since the network does not have any fully connected layers (which
consume $>$90\% of the parameters in a typical image processing CNN)
it is much lighter. \segnet is thus very memory-efficient and
comparatively easy to train.
We found, in agreement with its creators, that \segnet on its
own does not always reach the performance of much heavier architectures
with fully connected layers.
However, it turns out that \emph{in combination with learned class
  boundaries}, \segnet matches the more expensive competitors,
see Section~\ref{sec:experiments}.

Our variant, denotes as \segneth, consists of two parallel \segnet
branches, one for the colour channels and one for the digital
elevation model (DEM).
The colour branch is initialized with the existing \segnet weights, as
trained on the Pascal dataset%
\footnote{\scriptsize{\url{http://mi.eng.cam.ac.uk/~agk34/resources/SegNet/segnet_pascal.caffemodel}}} %
by \citet{badrinarayanan2017segnet}.
The second branch processes a two-channel image consisting of nDSM and
DSM, and is initialized randomly using "Xavier" weight initialization,
a technique designed to keep the gradient magnitude roughly the same
across layers~\citep{glorot2010}.
The outputs from the two streams are then concatenated, and fed
through a 1$\times$1 convolution that linearly combines the vector of
feature responses at each location into a score per class. Those class
scores are further converted to probabilities with a \textit{softmax}
layer.

\citet{Kokkinos2016pushing} reported significant quantitative
improvements by an explicit multi-scale architecture, which passes
down-scaled versions of the input image through identical copies of
the network and fuses the results.
Given the small size of \segneth we have also experimented with that
strategy, using three scales. We thus set up three copies of the
described two-stream \segneth with individual per-scale weights. Their
final predictions, after fusing the image and height streams, are
upsampled as needed with fractional stride convolution layers%
\footnote{Sometimes inaccurately called ``deconvolution''
  layers. Technically, these layers perform convolution, but sample
  the input feature maps from the previous layer with a
  stride $<$1. E.g., a stride of $\frac{1}{2}$ will double the
  resolution.} %
and fused before the final prediction of the class scores.
The multi-scale strategy only slightly improves the results by
$<$0.5\%, presumably because remote sensing images, taken in nadir
direction from distant viewpoints, exhibit only little perspective
effect and thus less scale variation (only due to actual scale
variability in metric object coordinates).
Still, the small improvement is consistent over all tiles of our
validation set, hence we include the multi-scale option in the
high-end variant of our system.

\subsubsection*{\hedh boundary-detection network}

We aim to include explicit boundary information in our processing,
thus we require an edge detector that can be integrated into the
labeling framework.  In a nutshell, \hed (Holistically-Nested Edge
Detection) is an multi-scale encoder-decoder CNN trained to output an
image of edge likelihoods.
An important feature of \hed is that it uses multi-scale prediction in
conjunction with
deep supervision~\citep{lee2015deeply}. That is, the
(rectified) output of the last convolution before each pooling layer
is read out of the network as prediction for that particular scale,
and supervised during training by an additional (Euclidean distance)
loss function. The multi-scale predictions are then combined into a
final boundary probability map.
Importantly, even though \hed has no explicit mechanism to enforce
closed contours, we find that by learning from global image context it
does tend to generate closed contours, which is important to support
our segmentation task.

For our purposes, we again add a second branch for the DSM, and modify
\hed to better take into account the uncertainty of boundary
annotations, by using a regression loss w.r.t.\ a continuous
``boundary score'' $\mathbf{y}$ rather than a hard classification loss.
The intuition is that the location of a class boundary has an inherent
uncertainty. We therefore prefer to embrace and model the uncertainty,
and predict the ``likelihood'' of the boundary being present at a
certain pixel. The advantage is best explained by looking at pixels
very close to the annotated boundary: using those pixels as negative
(i.e., background) training samples will generate label noise; using
them as positive samples contradicts the aim to predict precise
boundary locations; and excluding them from the training altogether
means depriving the network of training samples close to the boundary,
where they matter most.
For our task, the desired edges are the segment boundaries, i.e., the
label transitions in the training data. To turn them into soft scores
we dilate the binary boundary image with a diamond structuring
element. The structuring element defines the width of the
``uncertainty band'' around an annotated boundary and depends on the
ground sampling distance.

\begin{figure}[]
  \includegraphics[width=\textwidth]{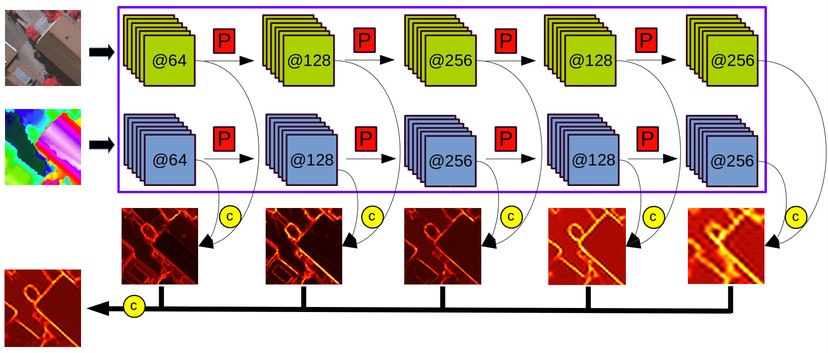}
  \caption{The Class-Boundary network \hedh. Color
    and height are processed in separate streams. Before each pooling
    level (red squares) the network outputs a set of scale-dependent
    class-boundaries, which are fused into the final multi-scale
    boundary prediction.  Yellow circles denote concatenation of
    feature maps.  }
        \label{fig:edge-network}
\end{figure}

\begin{figure}[]
  \includegraphics[width=\textwidth]{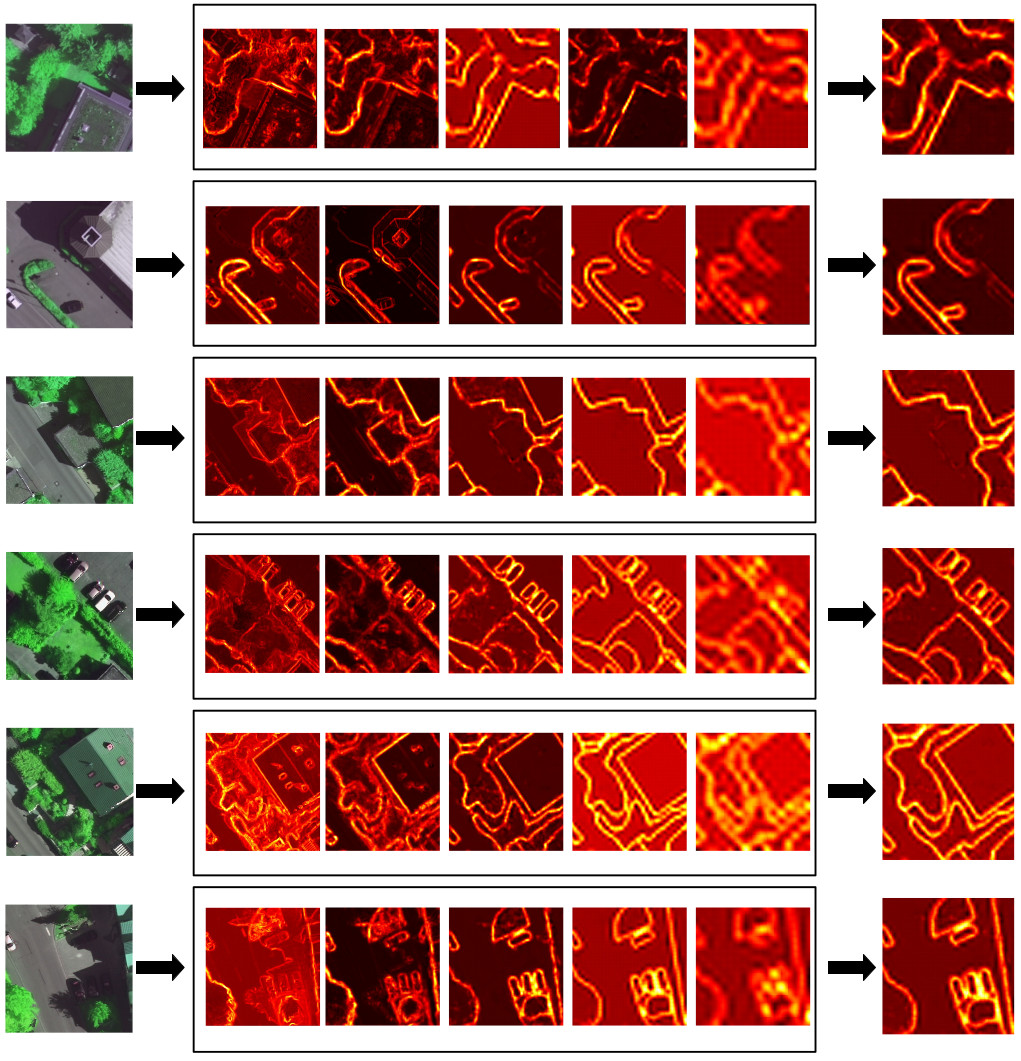}

    \caption{Examples of class boundary predictions. Left: input
      image. Middle: per-scale outputs. Right: estimated multi-scale
      boundary map.}
      	 \label{fig:class-contours-examples}
\end{figure}

Then, we weight each class-boundary pixel according
  to its truncated Euclidean distance to the nearest background pixel,
  using the distance transform: $Y=\beta\cdot D_{t}^{\ell
    2}(B_d)$.
The operator $D_{t}^{\ell 2}(B_d)$ denotes the truncated Euclidean
distance from a particular pixel to the nearest background pixel.
The factor $\beta=\frac{|B_d=0|}{|B_d|}$ compensates the relative
frequencies of boundary and background pixels, to avoid overfitting to
the dominant background.
Finally, the weights are normalized to the interval $[0\hdots 1]$
to be consistent with the original \hed model.
As above, we set up two separate streams for color images and for the
height also during boundary detection, and refer to our version as
\hedh. The image stream is initialized with the original \hed
weights, whereas the DEM stream is trained from scratch. For output
(including side outputs) the two streams are fused by concatenation,
convolution with a $1\times1$ kernel and fractional convolution to the
output resolution.
A graphical overview of the boundary network is given in
Figure~\ref{fig:edge-network}, visual results
of class-boundaries are shown in Figure~\ref{fig:class-contours-examples}.

\FloatBarrier

\subsubsection*{\dlreth semantic segmentation network}

From the literature~\citep{simonyan2014very} it is known that, also
for strong classifiers like CNNs, ensemble averaging tends to improve
classification results.
We thus also experiment with an ensemble of multiple CNNs.
To that end we use our previous semantic segmentation network, here
termed \dlreth, which has already shown competitive performance for
remote sensing problems~\citep{marmanis2016semantic}.
That model in fact is an ensemble of two identical \fcn architectures
initialized with different weights, namely those of \emph{VGG} and
\emph{Pascal}, see \cite{marmanis2016semantic}.
For clarity we refer to the ensemble as \dlreth, and to its two
members as \dlrethVGG, respectively \dlrethPAS.
Contrary to \segneth this model is derived from the standard \fcn and
has fully connected layers. It is thus heavier in terms of memory and
training, but empirically improves the predictions, presumably because
the network can learn extensive high-level, global object information.
Compared to the original \fcn~\citep{long2015fully}, our variant has
additional skip connections to minimize information loss from
downsampling. As above, it has separate streams for image and DEM
data.

\subsection{Integrated class boundary detection and segmentation}

\begin{figure}[]
  \centering
\includegraphics[width=0.8\textwidth]{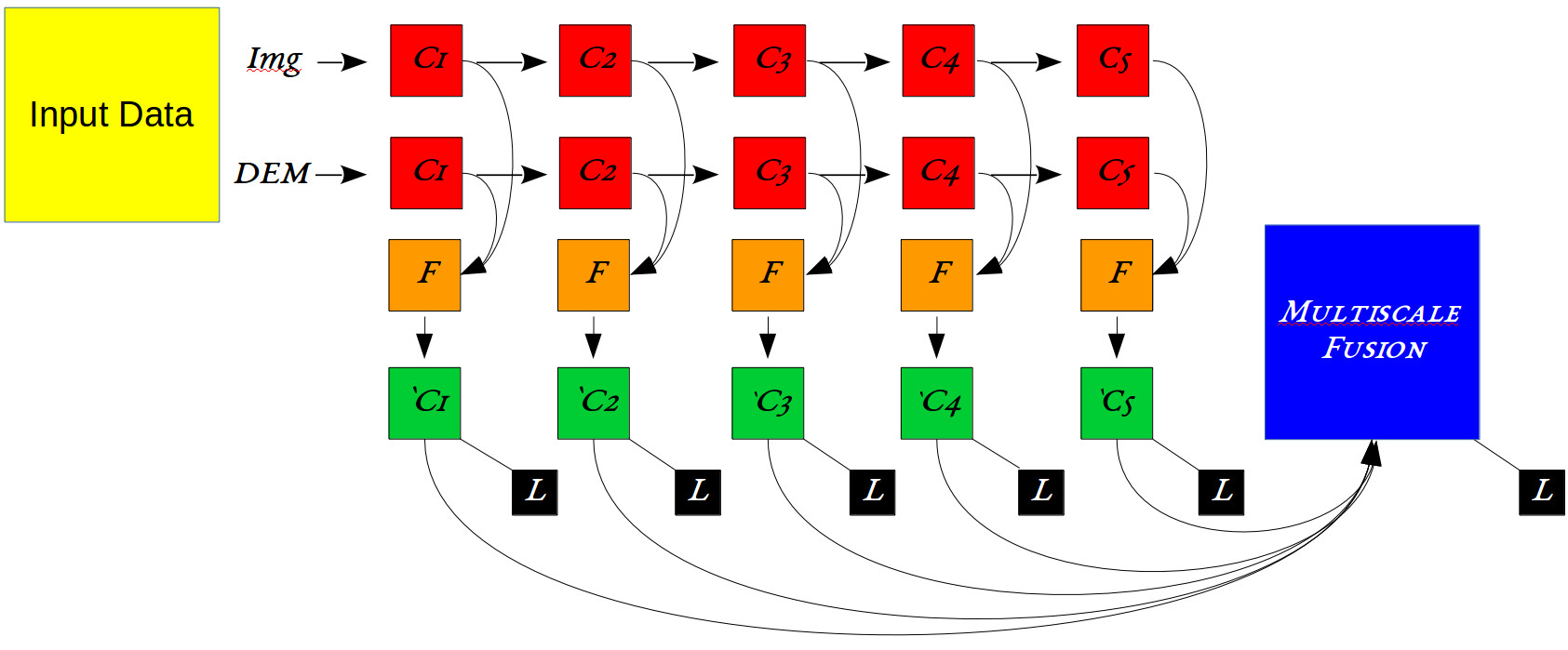}\\ [-0.5ex]
\includegraphics[trim={0 1cm 0 1.0cm},clip,width=\textwidth]{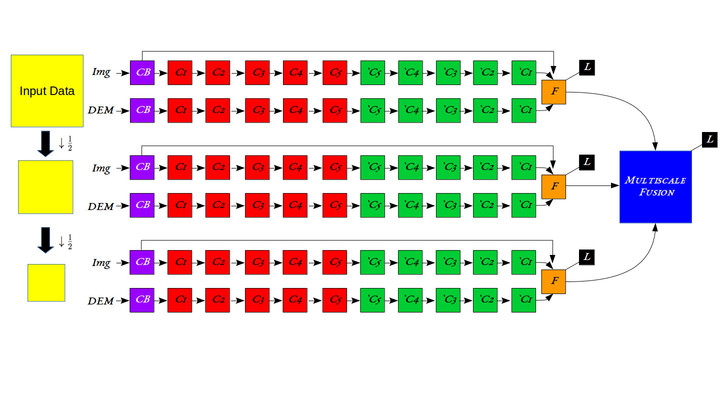}\\ [-2.5ex]
\includegraphics[trim={0 0 0 0.2cm},clip,width=\textwidth]{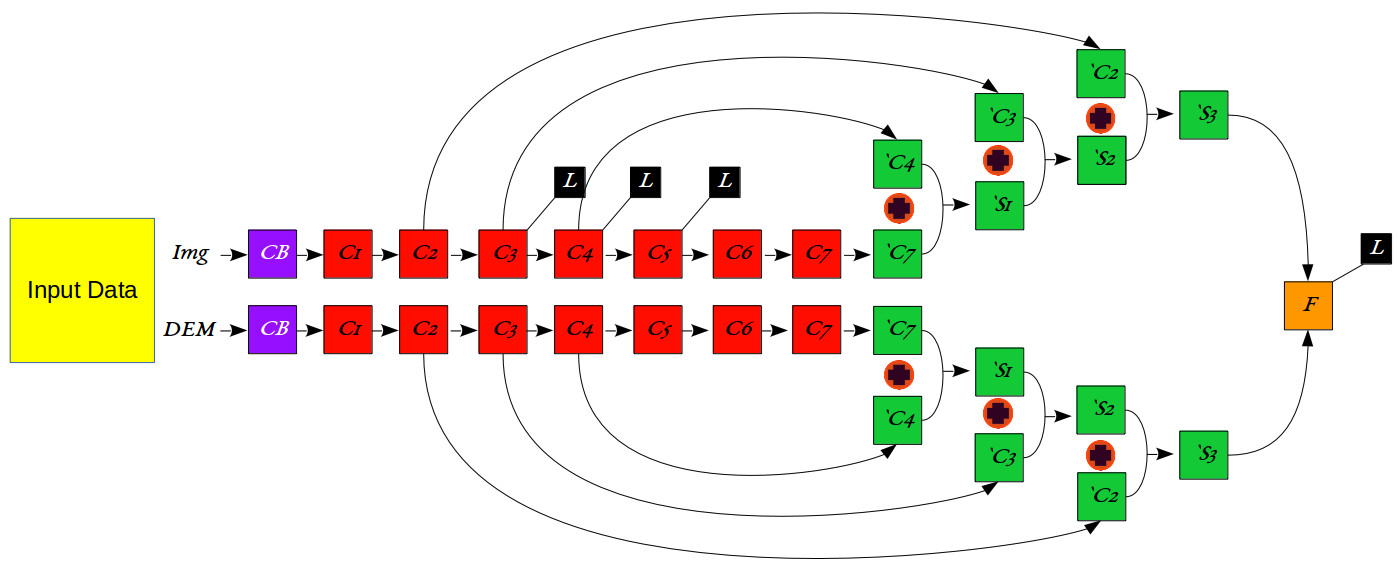}
\caption{CNN architectures tested in this work. Top:
  \hedh architecture for class-boundary delineation.
  Middle: multi-scale \segneth architecture.
  Bottom: \dlrethPAS architecture (identical to
  \dlrethVGG except for initial weights).  The
  boundary detection network is collapsed (violet box) for better
  readability. Encoder parts that reduce spatial resolution are marked
  in red, decoder parts that increase resolution are green. Orange
  denotes fusion by concatenation, $(1\times1)$ convolution, and
  upsampling (as required).  The red circle with the plus sign denotes
  element-wise summation of feature maps, the black box symbolises the
  (logistic, respectively Euclidean) loss.}
  \label{fig:full-architectures}
\end{figure}

Our strategy to combine class boundary detection with semantic
segmentation is straight-forward: we append the class-boundary network
(\hedh) before the segmentation network (\dlreth or \segneth) and
train the complete architecture.
In the image stream, the input is the raw colour image; in the DSM
stream it is a 2-channel image consisting of the raw DSM and the nDSM
(generated by automatically filtering above-ground objects and
subtracting the result from the DSM).
In both cases, the input is first passed through the \hedh network
(Figure~\ref{fig:full-architectures}(top)), producing a scalar image
of boundary likelihoods.
That image is concatenated to the raw input as an additional channel
to form the input for the corresponding stream of the semantic
segmentation network (note, in CNN language this can be seen as a skip
connection that bypasses the boundary detection layers).
For \segneth, a further skip connection bypasses most
of the segmentation network and reinjects the colour image boundaries
as an extra channel after merging the image and DSM streams, see
Figure~\ref{fig:full-architectures}(middle). This additional skip
connection re-introduces the class-boundaries deep into the
classifier, immediately before the final label prediction.
For \dlreth this did not seem necessary, since the
architecture already includes a number of long skip connections from
rather early layers, see Figure~\ref{fig:full-architectures}(bottom).
The entire processing chain is trained end-to-end (see below for
details), using boundaries as supervision signal for the \hedh layers
and segmentations for the remaining network.

\subsection{Ensemble learning}

As discussed above, ensemble averaging typically improves DCNN
predictions further.
We thus also test that strategy and combine the predictions of three
different boundary-aware networks (given the effort to train deep
networks, ``ensembles'' are typically small).
As described above, we have trained two versions of the \dlreth
network with integrated \hedh boundary detector, one initialized with
the original \fcn{} (Pascal VOC) weights%
\footnote{\scriptsize{\url{http://dl.caffe.berkeleyvision.org/fcn32s-heavy-pascal.caffemodel}}} %
and the other initialized with weights from the \vgg-16 variant.%
\footnote{\scriptsize{\url{http://www.robots.ox.ac.uk/~
    vgg/software/very\_deep/caffe/VGG\_ILSVRC\_16\_layers.caffemodel}}} %
Their individual performance is comparable, with \dlrethVGG
slightly better overall. In both cases, the DEM channel is again
trained from random initializations.

\begin{figure}[H]
\centering
  \includegraphics[trim={0 10cm 0 0},clip,width=0.85\textwidth]{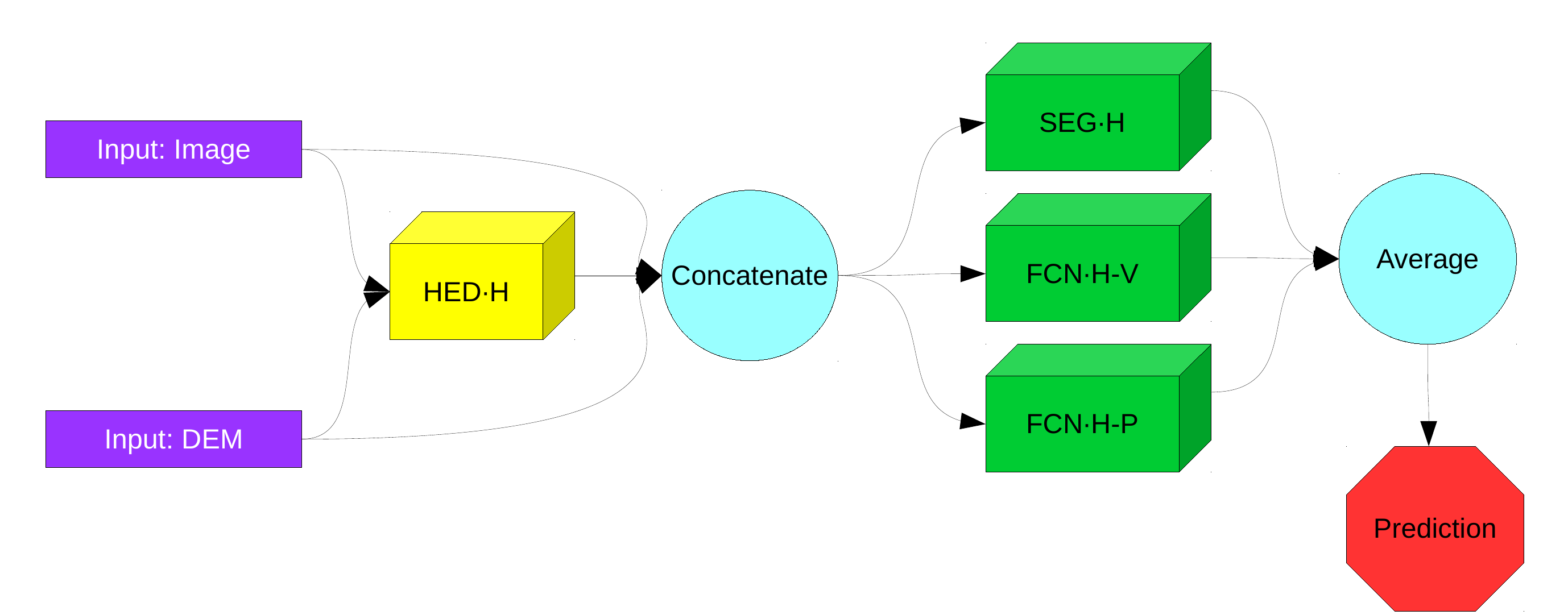}
\caption{Ensemble prediction with \segneth, \dlrethPAS and
    \dlrethVGG. The \hedh component extracts the class boundaries.}
\label{fig:ensemble-workflow}
\end{figure}

Ensemble predictions are made by simply averaging the
individual class probabilities predicted by \dlrethVGG, \dlrethPAS and \segneth.%
\footnote{We did not experiment with trained fusion layers, since the
  complete ensemble is too large to fit into GPU memory.} %
Empirically the ensemble predictions give a significant performance
boost, seemingly \dlreth and
\segneth are to some degree complementary,
see experimental results in Section~\ref{sec:experiments}.
Note though, the two \dlreth models, each with $>$100 million
parameters (see section \ref{sec:experiments}), are memory-hungry and
expensive to train, thus we do not generally recommend such a
procedure, except when aiming for highest accuracy.
Figure~\ref{fig:ensemble-workflow} depicts the complete ensemble.

\subsection{Implementation details and training}

The overall network with boundary detection is fairly deep, and the
boundary and labeling parts use different target outputs and loss
functions.
We found that under these circumstances training must proceed in
stages to achieve satisfactory performance, even if using pre-trained
components.
A cautious strategy, in which each component is first trained
separately, gave the best results. First, we train the boundary
detector (\hedh) separately, using \hed weights to
initialize the image stream and small random weights for the DSM
stream. That step yields a DCNN boundary detector tailored to our
aerial data.
The actual segmentation network and loss is added only after this
strong ``initialization'' of the boundary detector. Thus, the
boundary detector from the start delivers sensible results for
training the semantic labeling component, and only needs to be
fine-tuned to optimally interact with the subsequent layers.
Moreover, for \segneth that two-stage training is carried out
separately for each of the three scales. The separate single-scale
segmentation networks are then combined into one multi-scale
architecture and refined together.
Empirically, separating the scales stabilizes the learning of the
lower resolutions. When trained together immediately, they tend to
converge to weak solutions and the overall result is dominated by the
(still very competitive) highest resolution.

\paragraph{Normalization of gradients}
Regarding the common problem of exploding or vanishing gradients
during training, we stuck to the architectures recommended by the
original authors, meaning that \segneth does
use \emph{batch normalization} \citep{badrinarayanan2017segnet},
while \hedh does not \citep{xie2015holistically}.
A pragmatic solution is to use a large base learning rate appropriate
for \segneth and add layer-specific scale factors to decrease the
learning rate in the \hedh layers.
We also found that batch normalization in the final layers, after the
\segneth decoder phase, strongly sparsifies the feature
maps. For our goal of dense per-pixel prediction this effect is
detrimental, causing a $\approx\,$1\% drop in labeling accuracy.
We thus switch-off batch normalization for those layers.

\paragraph{Drop-out}
The original \segnet relies heavily on drop-out. The authors recommend
to randomly drop 50\% of the trainable decoder weights in each
iteration.
We found this drastic regularization to negatively affect our results,
thus we decrease it to 20\% for the highest resolution, respectively
10\% for the two lower ones.
Further research is needed to understand this big difference. We
suspect that it might have to do with the different image statistics.
In close-range images, each object normally covers a large image area,
thus both image intensities and semantic labels are strongly
correlated over fairly large regions.
In remote sensing data, with its many small objects, nearby
activations might not be able to ``stand in'' as easily for a dropped
connection, especially in the early layers of the decoder.

\paragraph{Data Augmentation}
DCNNs need large amounts of training data, which are not always
available.
It is standard practice to artificially increase the amount and
variation of training data by randomly applying plausible
transformations to the inputs.
Synthetic data augmentation is particularly relevant for remote
sensing data to avoid over-fitting: in a typical mapping project the
training data comes in the form of a few spatially contiguous regions
that have been annotated, not as individual pixels randomly scattered
across the region of interest. This means that the network is prone to
learn local biases like the size of houses or the orientation of the
road network particular to the training region.
Random transformations -- in the example, scaling and rotation -- will
mitigate such over-fitting.

In our experiments we used the following transformations for data
augmentation, randomly sampled per mini-batch of the stochastic
gradient descent (SGD) optimisation: scaling in the range $[1\hdots
  1.2]$, rotation by $[0^\circ\hdots 15^\circ]$ degrees, linear shear
with $[0^\circ\hdots 8^\circ]$, translation by $[-5\hdots 5]$ pixels,
and reflections w.r.t.\ the vertical and horizontal axis
(independently, with equal probability).

\section{Experiment Results}\label{sec:experiments}

\subsection{Datasets}
\bigskip
\subsection*{ISPRS Vaihingen Dataset}

We conduct experimental evaluations on the ISPRS Vaihingen 2D semantic
labeling challenge.
This is an open benchmark dataset provided online.%
\footnote{\scriptsize{\url{http://www2.isprs.org/commissions/comm3/wg4/2d-sem-label-vaihingen.html}}} %
The dataset consists of a color infrared orthophoto, a DSM
generated by dense image matching, and manually annotated ground truth
labels. Additionally, a nDSM has been released by one of the
organizers \citep{markususe}, generated by automatically
filtering the DSM to a DTM and subtracting the two.
Overall, there are $33$ tiles of $\approx2500\times2000$ pixels at a
GSD of $\approx9$cm.
$16$ tiles are available for training and validation, the remaining
$17$ are withheld by the challenge organizers for testing.
We thus remove $4$ tiles (image numbers 5, 7, 23, 30) from the
training data and use them as validation set for our experiments. All
results refer to that validation set, unless noted otherwise.


\begin{figure}[t!]
\centering
\begin{tabular}{lll}
  \includegraphics[width=0.3\textwidth]{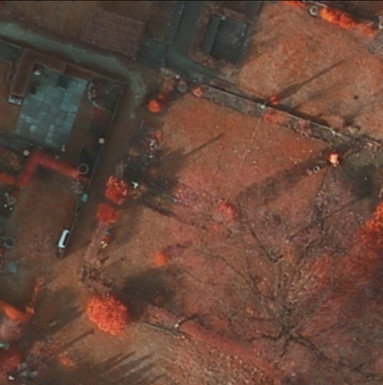} &
  \includegraphics[width=0.3\textwidth]{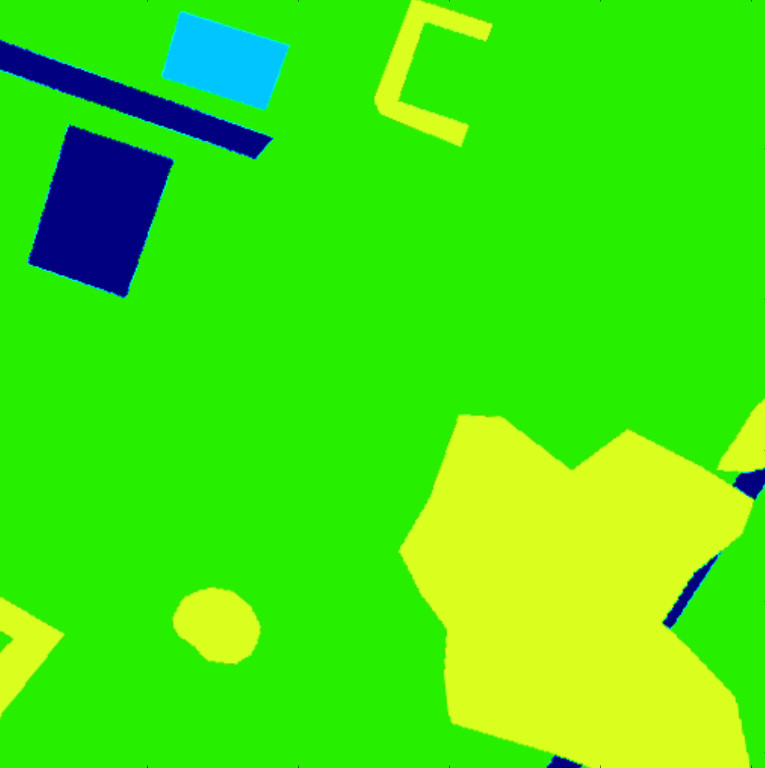} &
  \includegraphics[width=0.3\textwidth]{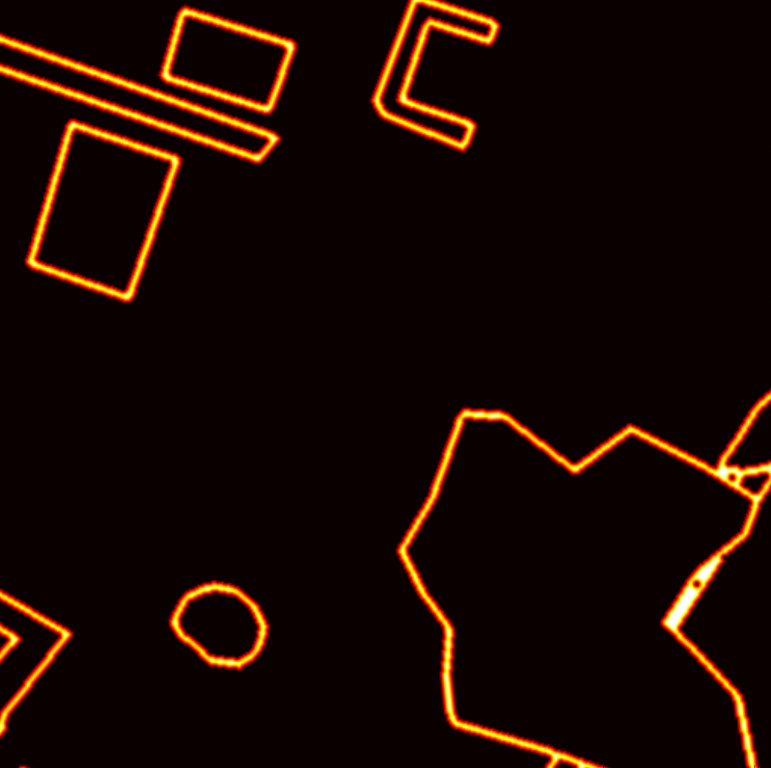}
\end{tabular}
\caption{Annotation ambiguity of trees in leaf-off condition in the
  Potsdam dataset.  Even though there is no image evidence (left),
  annotators tend to hallucinate a tree crown around the empty
  branches, denoted by yellow color in the labels (middle). The
  corresponding label class-boundaries (right) do not reflect
  discontinuities in the input images, thus contradicting our model
  assumptions.}
\label{fig:unreal-trees}
\end{figure}


\subsection*{ISPRS Potsdam Dataset}

We additionally conduct experiments on the ISPRS
  Potsdam semantic labeling dataset.%
  \footnote{\scriptsize{\url{http://www2.isprs.org/commissions/comm3/wg4/2d-sem-label-potsdam.html}}} %
  The data is rather similar to Vaihingen, with 4-band RGB-IR images,
  as well as DSM and nDSM from dense matching and filtering.
  There are $38$ image tiles of size $6000\times6000$ pixels,
with GSD $\approx5$cm.
$24$ tiles are densely annotated for training, whereas the remaining
14 are withheld as test set. For our experiments we have removed four
tiles (image numbers 7\_8, 4\_10, 2\_11, 5\_11) from the training data
and use them as validation set. All statistics on Potsdam refer to
that validation set.

We point out an issue with the reference data of Potsdam, which has
implications for our method: the trees are mostly deciduous and the
data was recorded in leaf-off conditions. However, the human
annotators appear to have guessed the area of the tree crown from the
branches and have drawn an imaginary, solid crown area. This
annotation contradicts the image evidence, see
Figure~\ref{fig:unreal-trees}. In particular, semantic boundaries
extracted from the reference data are not aligned with actual image
discontinuities.  We also note that, other than in Vaihingen, there
are comparatively large areas marked as \emph{background} (rejection)
class, with large intra-class variability.

\subsubsection*{Training details and parameters}

As described above, each labeling and boundary detection network was
first trained individually to convergence, then the pretrained pieces
were assembled and fine-tuned by further training
iterations. For a tally of the unknown weights,
see Table~\ref{tab:params}.

For the individual network parts, we always start from a quite high
learning rate (\textit{lr=$10^{-8}$}) and decrease it by a factor
$\times10$ every $12000$ iterations. The total of number of iterations
was $100000$. The \segneth part was trained with batch size $2$, the
\hedh boundary detector with batch size $5$.

The complete, assembled boundary+segmentation model was trained for
$30000$ iterations, starting with a smaller learning rate
(\textit{lr=$10^{-12}$}).  Batch size had to be reduced to $1$ to
stay within the memory capacity of an Nvidia Titan-X GPU.

The remaining hyper-parameters were set to
$momentum=0.9$ and $weight-decay=0.00015$, for all models.

\begin{table}[ht]
\centering
\caption{Sizes of model components in terms of trainable parameters.
All models are dimensioned to fit on a single Nvidia Titan-X GPU
(except for the ensemble, for which averaging is done on the CPU).
Suffix \textit{sc1} denotes a single-scale model using only the full
resolution, \textit{Msc} denotes the multi-scale model.}
\label{tab:params}
\begin{tabular}{|l|r|}
\hline
\hedh{}+\segneth-\textit{sc1} &   $88\cdot10^6$ \\ \hline\hline
\hedh{}+\segneth-\textit{Msc} &   $206\cdot10^6$ \\
\hedh{}+\dlrethPAS 	      &   $300\cdot10^6$ \\
\hedh{}+\dlrethVGG            &   $300\cdot10^6$ \\ \hline
\hedh{}+\dlreth{}+\segneth    &   $806\cdot10^6$ \\ \hline
\end{tabular}
\end{table}

\subsection{Results}

We evaluate the different components of our model by gradually adding
components. We start from the basic \segneth, augment
it with class boundaries, then with multi-scale processing. Finally,
we include it in a DCNN ensemble.
We will not separately discuss post-processing of the outputs with
explicit smoothness or context models like CRFs.
In our tests, post-processing with a fully connected CRF, as for
example in \citep{ChenPK0Y16,marmanis2016semantic}, did not
improve the results -- if anything, it degraded the performance on
small objects, especially the \emph{car} class. For completeness,
results of our networks with and without smoothing are available on
the Vaihingen benchmark site.
We conclude that our DCNNs already capture the context in large
context windows. Indeed, this is in our view a main reason for their
excellent performance.

\subsubsection*{Basic CNN Results}

\textbf{Vaihingen:}
The basic single-scale \segneth reaches $84.8$\% overall
accuracy over the validation set.
This is in line with other researchers' findings on the Vaihingen
benchmark: straight-forward adaptations of state-of-the-art DCNN
models from the computer vision literature typically reach
around $85$\%.
Our network performs particularly well on impervious ground and
buildings, whereas it is challenged by low vegetation, which is
frequently confused with the tree class. Detailed per-class results
are given in Table~\ref{tab:perclass}.

For comparison, we also run our earlier \dlreth
model~\cite{marmanis2016semantic}, i.e., an ensemble of only
\dlrethVGG and \dlrethPAS, without explicit class-boundary detection.
That model performs comparably, with $85.5$\% overall
accuracy. Interestingly, it is significantly better at classifying low
vegetation, and also beats \segneth on impervious surfaces and trees.
On the contrary, it delivers clearly worse segmentations of buildings.

\textbf{Potsdam:} On the Potsdam dataset the
\segneth-\textit{sc1} network performs
significantly better than the standard \dlrethPAS
and \dlrethVGG, with $84.9$\%, $80.9$\% and $81.4$\%, overall
accuracy respectively.
The reason for the relatively weak results of the two larger networks is
unclear, but we did not manage to improve them further. We will see below
that including boundary information closes the performance gap.
For detailed results refer to Table~\ref{tab:perclass}.

\subsubsection*{Effect of Class Boundaries}

\textbf{Vaihingen:}
We now go on to evaluate the main claim of our paper, that explicit
class boundary detection within the network improves segmentation
performance.
Adding the \hedh boundary detector to \segneth reaches $89.8$\%
overall accuracy (\hedh{}+\segneth-\textit{sc1}), a gain of more than
$5$ percent points, see Table~\ref{tab:perclass}.

The per-class results in Table~\ref{tab:perclass} reveal that class
boundaries significantly boost the performance of
\segneth for \emph{all} target classes, including the
vegetation classes that do not have sharp, unambiguous boundaries.
We suspect that in vegetation areas, where matching-based DSMs are
particularly inaccurate, even imprecise boundaries can play a
noticeable role in delimiting high from low vegetation.
Moreover, one might speculate that boundaries derived from semantic
segmentation maps carry information about the extent and
discontinuities at object level, and can to some degree mitigate a
main limitation of \segneth, namely the lack of fully
connected layers that could extract long-range context. It is however
an open, and rather far-reaching, question to what extent locally
derived object boundary information can substitute large-scale
semantic context.

For the \dlreth ensemble, we observe a similar, albeit
weaker effect. Overall accuracy increases by $3$ percent points to
$88.8$\% (\hedh{}+\dlreth). There are however losses
for the \emph{car} and \emph{low vegetation} classes, whereas the
gains are due to better segmentation of buildings and trees.

As a side note, we found that the accuracy of the ground truth is
noticeably lower for the vegetation classes as well as the cars.
This can in part be explained by inherent definition uncertainty,
especially in the case of vegetation.
Still, we claim that the upper performance bound for automatic
segmentation methods is probably well below $100$\% for Vaihingen, due
to uncertainty and biases of the ground truth of both the training and
test set.
See also Section~\ref{sec:experiments-manual}.

\textbf{Potsdam:} Similar effects can
be observed on the Potsdam dataset, where the class boundaries also
significantly improve the overall accuracy by up to $4.5$ percent
points. Detailed results are given in Table~\ref{tab:perclass}.
An exception from this trend is the tree class in
\segneth-\textit{sc1}: adding class boundaries reduces its
correctness significantly, from $74.4$\% to $68.6$\%.
We attribute this to the guessed tree-crown annotations mentioned
above. Since their imaginary boundaries are not reflected in the image
data, adding them apparently hurts, rather than helps the localization
of the tree outlines.
Curiously, the trees do get a boost from boundary information when
using the \dlrethPAS or \dlrethVGG networks. We were not yet able to
determine the reason for this behavior. We speculate that possibly
these networks, with their higher capacity and pre-training for object
detection, can extract additional evidence for the presence/absence of
a tree from the edge maps and are better able to infer the ``guessed''
tree outline from context.

\begin{table}[ht]
\centering
\caption{Adding an explicit class-boundary model improves semantic
  segmentation, for both tested CNN architectures and both
  investigated datasets.  \hedh and \emph{sc1} denote use of
  class-boundaries and restriction to a single scale,
  respectively. Scores are (true positive) detection rates. See text
  for details.}
\label{tab:perclass}
\begin{adjustbox}{center}
  \small
\begin{tabular}{|c|l|ccccc||c|}
\hline
& & \textit{Impervious} & \textit{Building} & \textit{Low Veg.} & \textit{Tree} & \textit{Car} & \textit{OA} \\ \hline\hline
\multirow{4}{*}{Vaihingen}
& \segneth-\textit{sc1} &
87.5 \%  & 93.8 \%  & 59.0 \% & 79.0 \% & 63.0 \% &  84.8 \% \\
& \hedh{}+\segneth-\textit{sc1} &
91.2 \% & 95.6 \% & 70.8 \% & 92.3 \% & 69.0 \% &  \textbf{89.8} \% \\ \cline{2-8}
& \dlreth &
89.3 \% & 87.5 \% & 77.3 \% & 88.8 \% & 68.3 \% & 85.8 \% \\
& \hedh{}+\dlreth &
89.3 \% & 93.5 \% & 73.0 \% & 90.8 \% & 62.0 \% & 88.8 \% \\ \hline\hline
\multirow{6}{*}{Potsdam}
& \segneth-\textit{sc1} &
84.7 \% & 95.2 \% & 78.9 \% & 74.4 \% & 80.8 \% &  84.9 \% \\
& \hedh{}+\segneth-\textit{sc1} &
85.0 \% & 96.7 \% & 84.2 \% & 68.6 \% & 85.8 \% & \textbf{85.1} \% \\ \cline{2-8}
& \dlrethPAS &
84.0 \% & 88.5 \% & 72.0 \% & 74.7 \% & 75.0 \% & 80.9 \% \\
& \hedh{}+\dlrethPAS &
84.9 \% & 96.2 \% & 76.7 \% & 79.1 \% & 86.7 \% & 84.5 \% \\ \cline{2-8}
& \dlrethVGG &
84.3 \% & 90.6 \% & 71.8 \% & 74.9 \% & 74.4 \% & 81.4 \% \\ 
& \hedh{}+\dlrethVGG &
85.2 \% & 96.8 \% & 74.3 \% & 79.1 \% & 85.4 \% & 85.0 \% \\ \hline
\end{tabular}
\end{adjustbox}
\end{table}

\subsubsection*{Effect of Multi-scale CNN}

\textbf{Vaihingen:}
Next, we test what can be gained by explicit multi-scale
processing.
This is inspired by \citet{Kokkinos2016pushing}, who show significant
improvements with multi-scale processing.
Our implementation uses exactly the same architecture, with three
streams that independently process inputs of different
scale and fuse the results within the network.

We run this option only for \segneth. The \dlreth network has
multiple fully connected layers and should therefore be able to
capture context globally over the entire input region, thus we do not
expect an explicit multi-scale version to improve the
results. Moreover, it is too large to fit multiple copies of the
network into GPU memory.

Empirically, multi-scale processing did not improve the results to the
same extent as in \citep{Kokkinos2016pushing}. We only gain $0.2$
percent points, see Table~\ref{tab:multiscale}. Apparently, the
single-scale network already captures the relevant information. We
suspect that the gains from an explicit multi-scale architecture are
largely achieved by better covering strong scale variations due to
different perspective effects and camera viewpoints.
In nadir-looking remote sensing images such effects are absent and
scale variations occur only due to actual size variations in object
space, which seem to be adequately captured by the simpler network.
Nevertheless, since the gains are quite consistent across different
validation images, we keep the multi-scale option included for the
remaining tests on Vaihingen. We note though, this triples the memory
consumption and is therefore not generally recommended.
Given the small differences, we did not employ
multi-scale processing in the Potsdam experiments.

\begin{table}[ht]
  \caption{Multi-scale processing and ensemble learning over the ISPRS
    Vaihingen dataset. The results (overall accuracies) on the
    validation set confirm that gains are mostly due to the class
    boundary detection, whereas multi-scale processing and ensemble
    prediction only slightly improve the results further. See text for
    details.}
\label{tab:multiscale}
{\footnotesize
\begin{adjustbox}{center}
\begin{tabular}{|c|c|c|c|c|c|c|c|}
\hline
& \textit{Scene}
& \hedh{}+\dlreth{}+\segneth
& \hedh{}+\segneth-\textit{Msc}
& \hedh{}+\segneth-\textit{sc1}
& \segneth-\textit{sc1}
& \hedh{}+\dlreth
& \dlreth \\
\hline
\multirow{5}{*}{\rotatebox{90}{Vaihingen}}
&\textit{Image 5} &
91.5 \% & 91.2 \% & 91.1 \% & 86.2 \% & 90.4 \% & 86.3 \% \\
&\textit{Image 7} &
89.2 \% & 89.6 \% & 89.6 \% & 84.2 \% & 89.1 \% & 87.1 \% \\
&\textit{Image 23} &
92.0 \% & 90.8 \% & 90.2 \% & 85.6 \% & 89.3 \% & 83.7 \% \\
&\textit{Image 30} &
88.7 \% & 88.5 \% & 88.4 \% & 83.2 \% & 86.7 \% & 86.1 \% \\ \cline{2-8}
&\textit{OA} &
\textbf{90.3 \%} & 90.0 \% & 89.8 \% & 84.8 \% & 88.8 \% & 85.8 \% \\
\hline
\end{tabular}
\end{adjustbox}
}
\end{table}


\begin{table}[ht]
  \caption{Results on Potsdam validation set for individual networks
    and network ensemble. The class boundaries improve the
    classification accuracy of all three individual networks, and thus
    also of the ensemble.}
    \label{tab:Postdam-validation}
      {\footnotesize
    \begin{adjustbox}{center}
\begin{tabular}{|c|c|c|c|c|c|c|c|c|}
\hline & \textit{Scene} & \hedh{}+\dlreth{}+\segneth & \hedh{}+\segneth-\textit{sc1} & \segneth-\textit{sc1} & \hedh{}+\dlrethPAS & \dlrethPAS & \hedh{}+\dlrethVGG & \dlrethVGG \\ \hline
\multirow{5}{*}{\rotatebox{90}{Potsdam}}
&\textit{Image 2\_11} &
84.5 \% & 81.1 \% & 81.9 \% & 82.2 \% & 76.2 \% & 82.9 \% & 78.8 \% \\
&\textit{Image 5\_11} &
90.9 \% & 89.8 \% & 89.3 \% & 90.2 \% & 86.2 \% & 90.6 \% & 86.7 \% \\
&\textit{Image 4\_10} &
81.8 \% & 83.3 \% & 82.7 \% & 79.2 \% & 79.2 \% & 79.6 \% & 79.2 \% \\
&\textit{Image 7\_8 } &
87.7 \% & 86.3 \% & 85.6 \% & 86.6 \% & 81.6 \% & 86.7 \% & 80.9 \% \\
\cline{2-9}
&\textit{OA} &
\textbf{86.2 \%} & 85.1 \% & 84.9 \% & 84.5 \% & 80.9 \% & 85.0 \% & 81.4 \% \\
\hline
\end{tabular}
\end{adjustbox}
}
\end{table}

\subsubsection*{Effect of the Ensemble}

\textbf{Vaihingen:}
Several works have confirmed that also for DCNN models ensemble
learning is beneficial to reduce individual model biases. We have also
observed this effect in earlier work and thus test what can be gained
by combining several \emph{boundary-aware} segmentation networks.

We run the three introduced networks \segneth, \dlrethPAS and
\dlrethVGG, all with an integrated \hedh boundary detector, and
average their predictions.
The ensemble beats both the stand-alone \segneth model
and the two-model ensemble of (boundary-enhanced) \dlreth,
see Table~\ref{tab:multiscale}.
The advantage over \segneth
is marginal, whereas a clear improvement can be observed over \dlreth.
In other words, \segneth alone stays behind its
\dlrethVGG and \dlrethPAS
counterparts, but when augmented with class-boundaries it outperforms
them, and reaches almost the performance of the full ensemble. It
seems that for the lighter and less global \segneth model the boundary
information is particularly helpful.

\begin{figure}[htbp]
\centering
\begin{adjustbox}{center}
\begin{tabular}{cccc}
image & \dlreth & \hedh{}+\segneth & \hedh{}+\dlreth{}+\segnet \\
\includegraphics[width=0.24\textwidth]{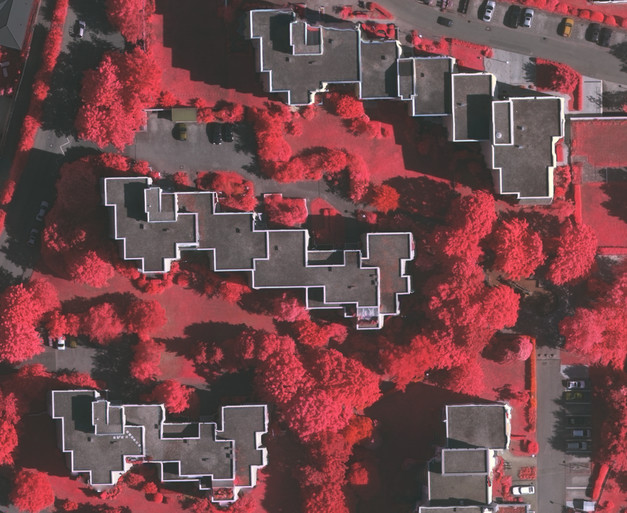} &
\includegraphics[width=0.24\textwidth]{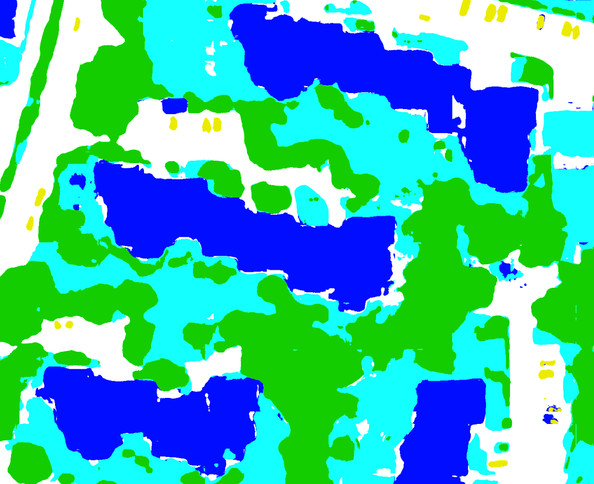} &
\includegraphics[width=0.24\textwidth]{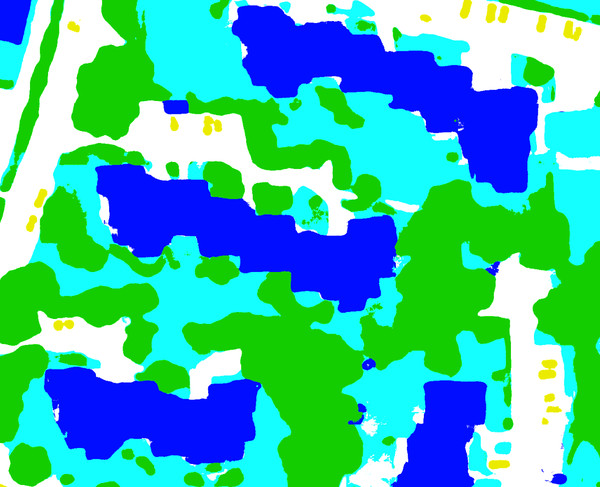} &
\includegraphics[width=0.24\textwidth]{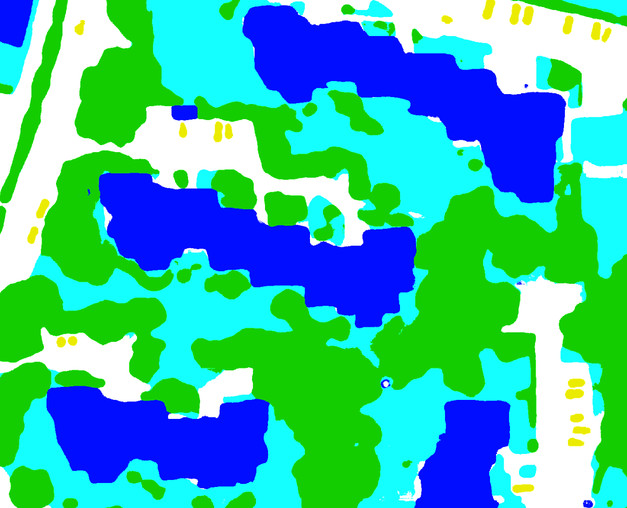} \\
& & & \\
\includegraphics[angle=90,origin=c,width=0.24\textwidth]{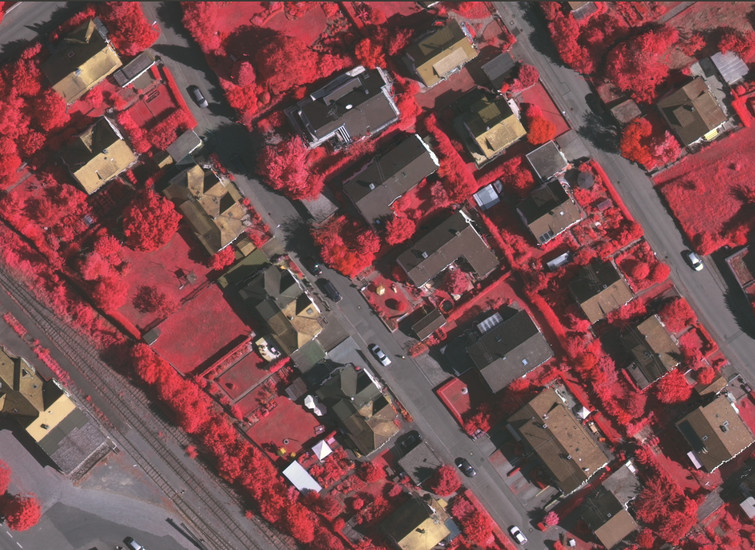} &
\includegraphics[angle=90,origin=c,width=0.24\textwidth]{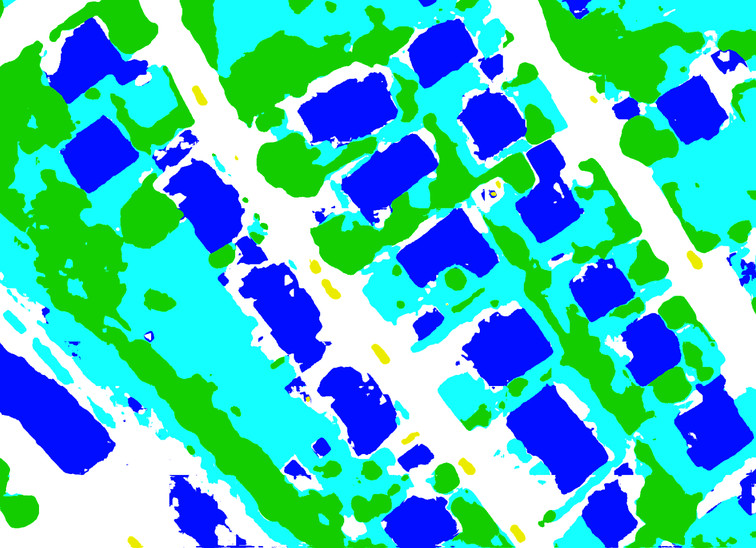} &
\includegraphics[angle=90,origin=c,width=0.24\textwidth]{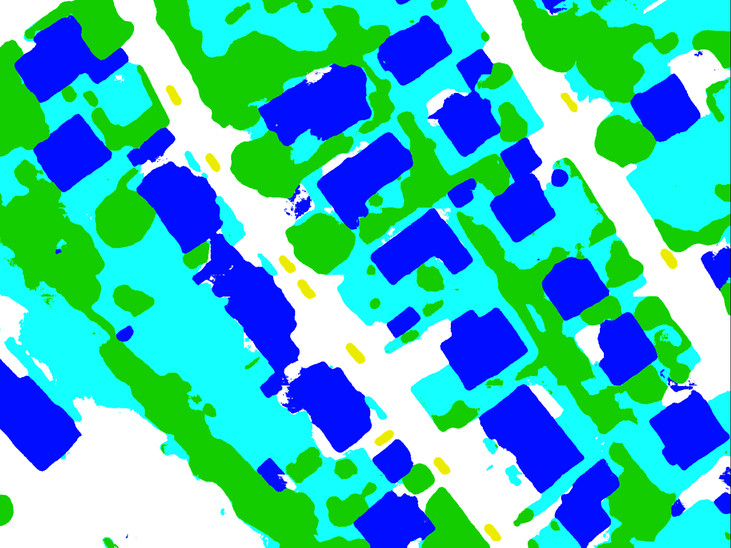} &
\includegraphics[angle=90,origin=c,width=0.24\textwidth]{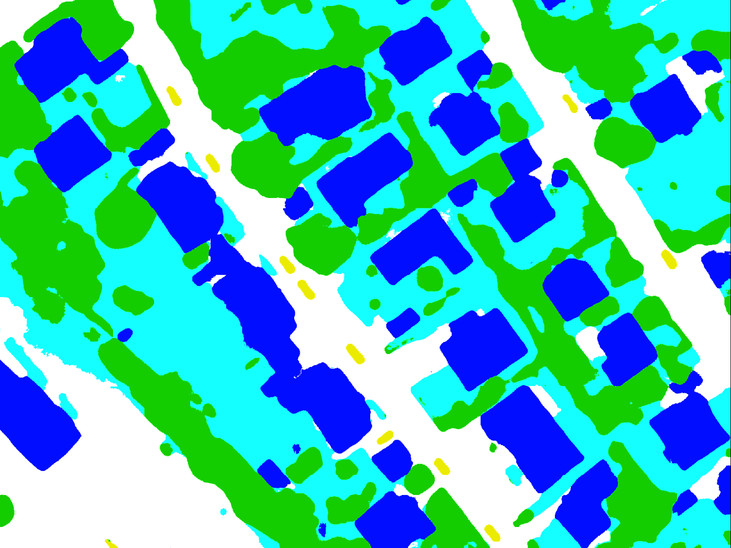} \\
& & & \\
\includegraphics[width=0.24\textwidth]{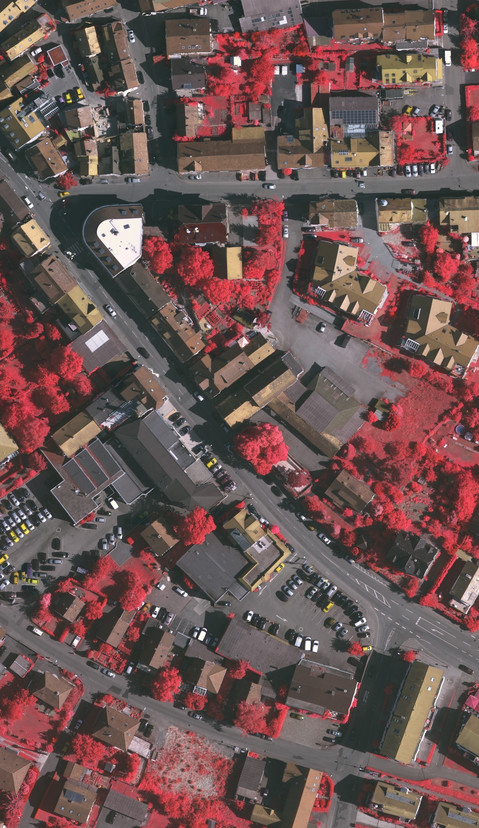} &
\includegraphics[width=0.24\textwidth]{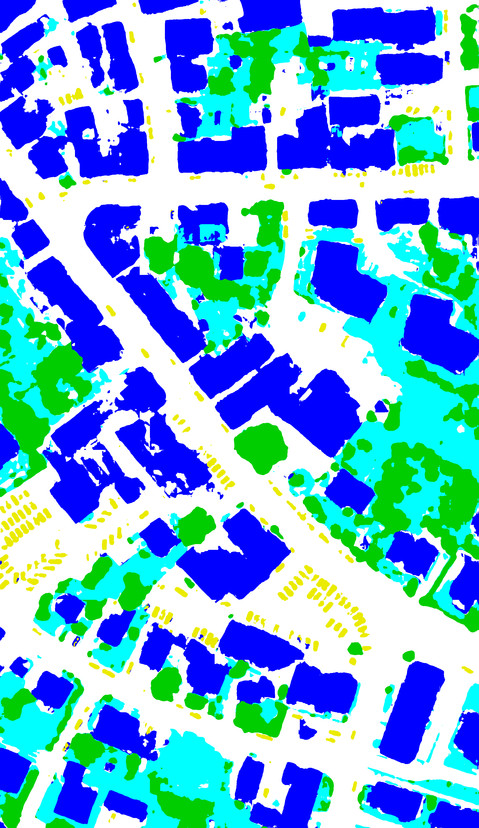} &
\includegraphics[width=0.24\textwidth]{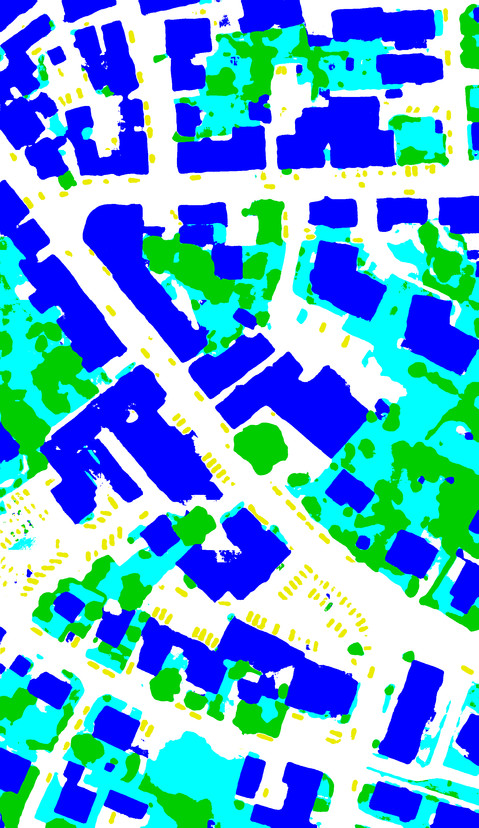} &
\includegraphics[width=0.24\textwidth]{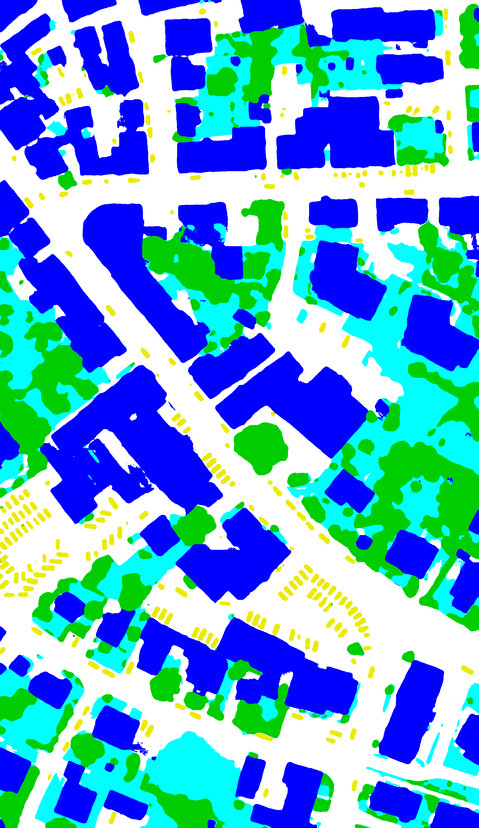} \\
\end{tabular}
\end{adjustbox}
\caption{Example predictions on the official Vaihingen test set. The
second to fourth column show the outputs of \dlreth (DLR\_1),
  \hedh+\segnet (DLR\_5) and \hedh+\dlreth+\segnet (DLR\_7).
  \textbf{White}: Impervious Surfaces, \textbf{Blue}: Buildings,
  \textbf{Cyan}: Low-Vegetation, \textbf{Green}: Trees,
  \textbf{Yellow}: Cars.}
\label{fig:Vaihingen-visualization}
\end{figure}

We point out that, by and large, the quantitative behavior is
consistent across the four individual tiles of the validation set. In
all four cases, \hedh{}+\dlreth clearly beats \dlreth, and similarly
\hedh{}+\segneth-\textit{sc1} comprehensively beats
\segneth-\textit{sc1}.
Regarding multi-scale processing, \hedh{}+\segneth-\textit{Msc} wins
over \hedh{}+\segneth-\textit{sc1} except for one case (image 7), where the
difference are a barely noticeable $0.03$ percent points (ca.\ 1500
pixels / 12 m$^\text{2}$).
Ensemble averaging again helps for the other three test tiles, with an
average gain of $0.21$ percent points, while for image 7 the ensemble
prediction is better than \hedh{}+\dlreth but does not quite reach
the \segneth performance.
A further analysis of the results is left for future work, but will
likely require a larger test set.
For a visual impression, see
Figure~\ref{fig:Vaihingen-visualization}.

%

\begin{figure}[ht]
\centering
\begin{adjustbox}{center}
\begin{tabular}{cccc}
image & \segneth & \hedh{}+\segneth & \hedh{}+\dlreth{}+\segneth \\
\includegraphics[angle=90,origin=c,width=0.24\textwidth]{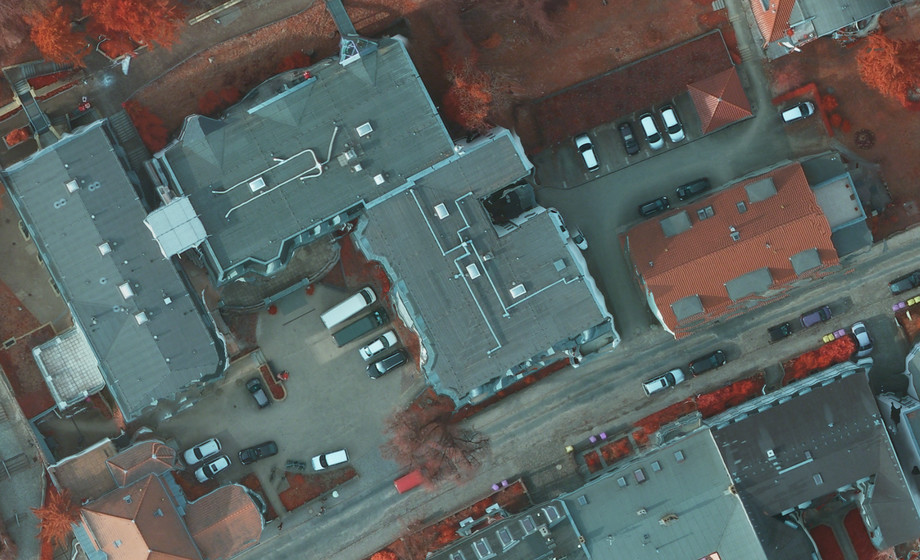} &
\includegraphics[angle=90,origin=c,width=0.24\textwidth]{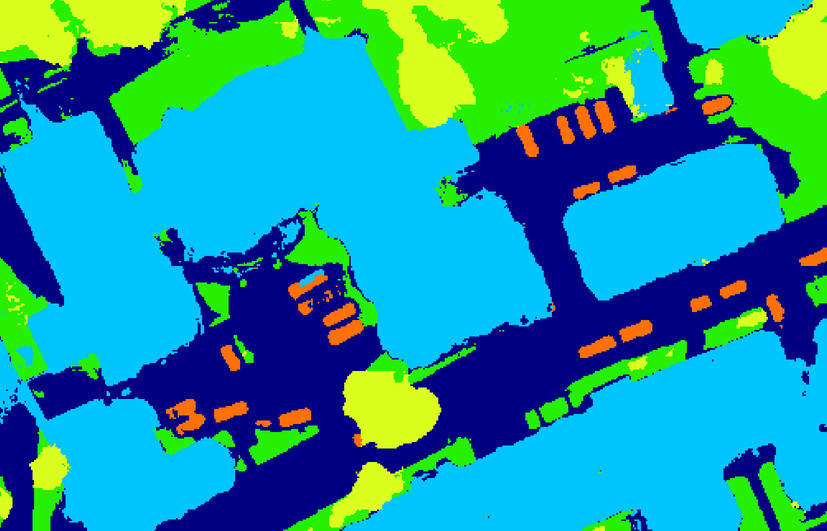} &
\includegraphics[angle=90,origin=c,width=0.24\textwidth]{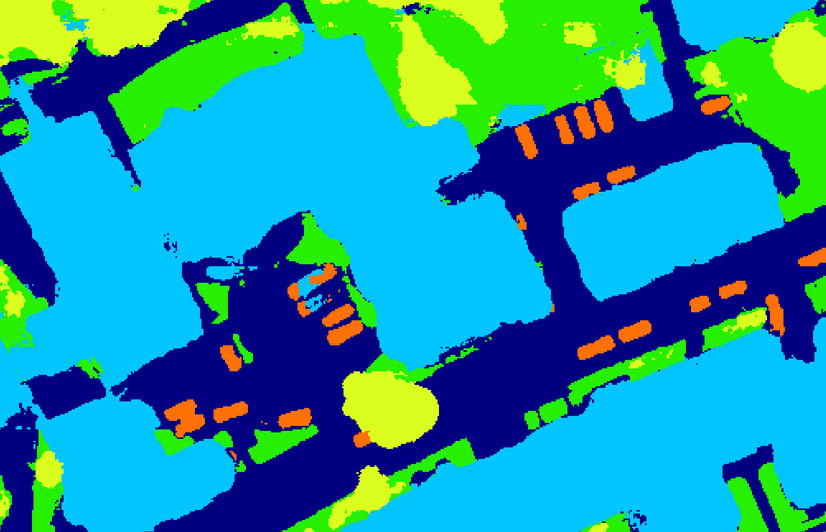} &
\includegraphics[angle=90,origin=c,width=0.24\textwidth]{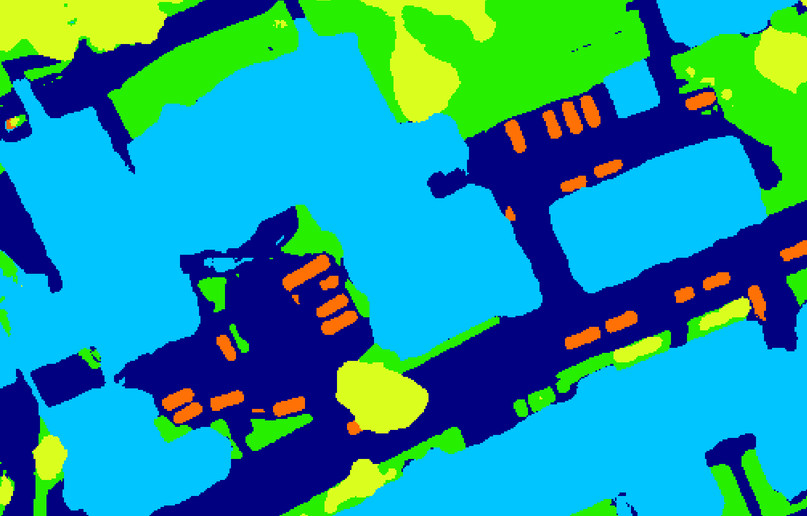} \\
& & & \\
\includegraphics[angle=90,origin=c,width=0.24\textwidth]{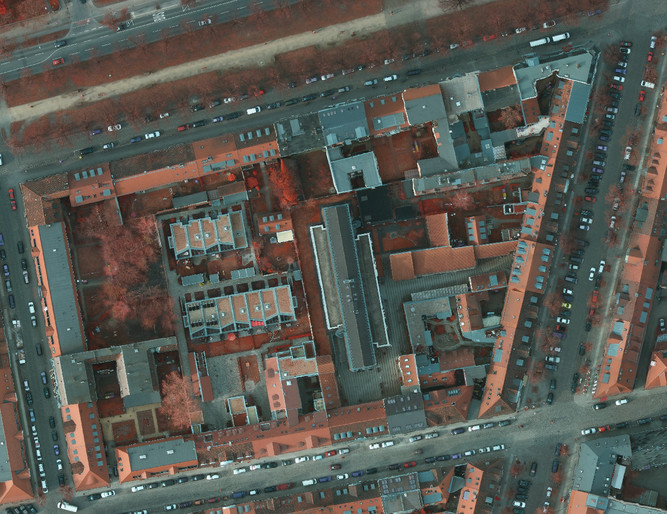} &
\includegraphics[angle=90,origin=c,width=0.24\textwidth]{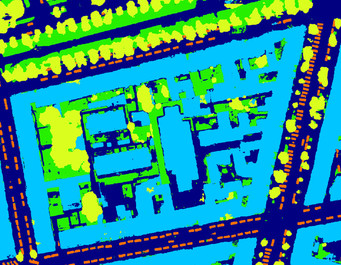} &
\includegraphics[angle=90,origin=c,width=0.24\textwidth]{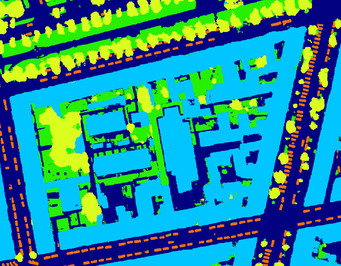} &
\includegraphics[angle=90,origin=c,width=0.24\textwidth]{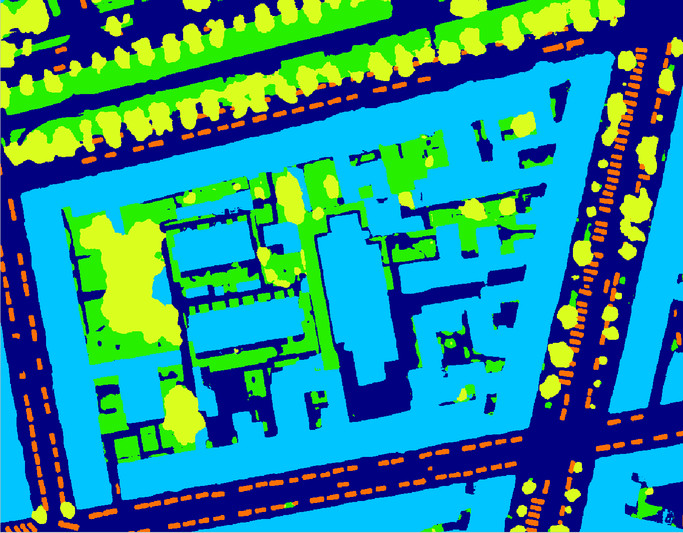} \\
\end{tabular}
\end{adjustbox}
\caption{Example predictions on our Potsdam validation set.  The
  second to fourth columns show results for \segneth, \hedh{}+\segneth
  and \hedh{}+\dlreth{}+\segneth.  \textbf{Dark Blue}: Impervious
  Surfaces, \textbf{Light Blue}: Buildings, \textbf{Green}:
  Low-Vegetation, \textbf{Yellow}: Trees, \textbf{Orange}: Cars.}
\label{fig:Potsdam-visualization}
\end{figure}

\textbf{Potsdam}:
For the ensemble model we proceed in the same way and average the
individual predictions of \segneth,
\dlrethPAS and \dlrethVGG.
Also for Potsdam, the ensemble with class boundary support for each of
the three members performs best, see
Table~\ref{tab:Postdam-validation}.
Note that, although all three networks exhibit almost identical
overall performance of $85\%$, averaging them boosts the accuracy by
another percent point.
Visual examples are shown in Figure~\ref{fig:Potsdam-visualization}.

\subsubsection*{Effects of nDSM Errors}

\textbf{Vaihingen:}
On the official Vaihingen test set the performance of our ensemble
drops to 89.4\%, see below. We have visually checked the results and
found a number of regions with large, uncharacteristic prediction
errors.
It turns out that there are gross errors in the test set that pose an
additional, presumably unintended, difficulty. In the nDSM
of~\citet{markususe}, a number of large industrial buildings are
missing, since the ``ground'' surface follows their roofs, most likely
due to incorrect filtering parameters.
The affected buildings cover a significant area: 3.1\% (154'752
pixels) of image 12, 9.3\% (471'686 pixels) of image 31, and 10.0\%
(403'818 pixels) of image 33.

By itself this systematic error could be regarded as a recurring
deficiency of automatically found nDSMs, which should be handled by the
semantic segmentation. But unfortunately, they only occur
in the test set, while in the training set no comparable
situations exist. It is thus impossible for a machine learning model
to handle them correctly.

To get an unbiased result we thus manually corrected the nDTMs of the
four affected buildings.
We then reran the testing, without altering the trained model in any
way, and obtained an overall accuracy of $90.3$\%, almost perfectly in
line with the one on our validation set, and $0.9$ percent points up
from the biased result.

In the following evaluation we thus quote both numbers. We regard the
$90.3$\% as our ``true'' performance on a test
set whose properties are adequately represented in the training data.
Since competing methods did however, to our knowledge, not use the
corrected test set, they should be compared to the $89.4$\% achieved
with the biased nDSM.
We note however that the discovered errors are significant in the
context of the benchmark: the bias of almost $1$ percent point is
larger than the typical differences between recent competing methods.
Since the experiment mainly serves to illustrate the quality and
influence of the available reference data, we did not repeat it for
Potsdam (where it would have been very hard to repair especially
the tree annotations).

\subsection{Comparison to state of the art}

\textbf{Vaihingen:}
Our proposed \textit{class-contour ensemble model} is among the top
performers on the official benchmark test set, reaching $89.4$\%
overall accuracy, respectively $90.3$\% with the correct nDSM. Note,
the model names on the benchmark website differ from those used here,
please refer to Table~\ref{tab:online-submitted-models}. The strongest
competitors at the time of writing%
  \footnote{The field moves forward at an astonishing pace, during the
    review of the present paper several groups have reached
    similar or even slightly higher accuracy. We cannot comment on
    these works, since no descriptions of their methodologies have
    appeared yet.} %
  were INRIA
\citep[$89.5$\%,][]{maggiori2016high}, using a variant of
\textsc{fcn}, and ONERA \citep[$89.8$\%,][]{audebert2016semantic},
with a variant of \segnet.
Importantly, we achieve above $90$\% accuracy over man-made classes,
which are the most well-defined ones, where accurate segmentation
boundaries matter most, see Table~\ref{tab:error-matrix}.

Detailed results for the top-ranking published models in the benchmark
are given in
Table~\ref{tab:compare-participants}. Table~\ref{tab:online-submitted-models}
  lists the benchmark identifiers of the different model variants we
  have described.
Overall, the performance of different models is very similar, which
may not be all that surprising, since the top performers are in fact
all variants of \fcn or \segnet.
We note that our model and INRIA are the most ``puristic'' ones, in that
they do not use any hand-engineered image features. ONERA uses the
NDVI as additional input channel; DST seemingly includes a random
forest in its ensemble, whose input features include the NDVI
(Normalized Vegetation Index) as well
as statistics over the DSM normals. It appears that the additional
features enable a similar performance boost as our class boundaries,
although it is at this point unclear whether they encode the same
information.
Interestingly, our model scores rather well on the \emph{car} class,
although we do not use any stratified sampling to boost rare
classes. We believe that this is in part a consequence of not
smoothing the label probabilities.

One can also see that after correcting the nDSM for large errors, our
performance is better than most competitors on impervious surfaces as
well as on buildings.
The bias in the test data thus seems to affect all models.
Somewhat surprisingly, our scores on the vegetation classes are also
on par with the competitors, although intuitively contours cannot
contribute as much for those classes, because their boundaries are not
well-defined. Still, they significantly improve the segmentation of
vegetation, c.f.~Table~\ref{tab:perclass}.
Empirically, the class-boundary information boosts segmentation of the
\emph{tree} and \emph{low vegetation} classes to a level reached by
models that use a dedicated NDVI channel. A closer look at the
underlying mechanisms is left for future work.

\begin{table}[ht]
\centering
\caption{Confusion matrix of our best result (\textit{DLR\_9}) on the
  private Vaihingen test set. Values are percent of predicted pixels
  (rows sum to 100\%, off-diagonal elements are false alarm rates).}
\label{tab:error-matrix}
{\small
\begin{tabular}{|cl|ccccc|}
\cline{3-7}
\multicolumn{2}{c|}{} & \multicolumn{5}{c|}{reference}\\
\multicolumn{2}{c|}{} & \textit{Impervious} & \textit{Building} & \textit{Low-Veg} & \textit{Tree} & \textit{Car} \\
\hline
& \textit{Impervious} &
\textbf{93.2 \%} & 2.2 \% & 3.6 \% & 0.9 \% & 0.1 \% \\
& \textit{Building} &
2.8 \% & \textbf{95.3 \%} & 1.6 \% & 0.3 \% & 0.0 \% \\
\parbox[t]{3mm}{\multirow{-2}{*}{\rotatebox[origin=c]{90}{predicted}}}
& \textit{Low-Veg} &
3.7 \% & 1.4 \% & \textbf{82.5 \%} & 12.5 \% & 0.0 \% \\
& \textit{Tree} &
0.7 \% & 0.2 \% & 6.9 \% & \textbf{92.3\%} & 0.0 \% \\
& \textit{Car} &
19.5 \% & 7.0 \% & 0.5 \% & 0.4 \% & \textbf{72.60 \%} \\
\hline
& \textit{Precision} &
91.6 \% & 95.0 \% & 85.5 \% & 87.5 \% & 92.1 \% \\
& \textit{Recall} &
93.2 \% & 95.3 \% & 82.5 \% & 92.3\% & 72.6 \% \\
& \textit{F1-score} &
92.4 \% & 95.2 \% & 83.9 \% & 89.9 \% & 81.2 \% \\
\hline
\end{tabular}
}
\end{table}

\begin{table}[th]
\centering
\caption{Short names of our model variants on the ISPRS Vaihingen 2D
  benchmark website.}
\label{tab:online-submitted-models}
\small
\begin{tabular}{|c|l|}
\hline
\textit{Abbreviation} & \multicolumn{1}{c|}{\textit{Model Details}} 		\\ \hline
DLR\_1  & \dlreth                                      \\
DLR\_2  & \dlreth{}+CRF                                \\
DLR\_3  & \hedh{}+\dlreth                              \\
DLR\_4  & \hedh{}+\dlreth{}+CRF                        \\
DLR\_5  & \hedh{}+\segnet                              \\
DLR\_6  & \hedh{}+\segneth{}+CRF                       \\
DLR\_7  & \hedh{}+\dlreth{}+\segneth                   \\
DLR\_8  & \hedh{}+\dlreth{}+\segneth{}+CRF             \\
DLR\_9  & \hedh{}+\dlreth{}+\segneth, nDSM corrections \\
DLR\_10 & \hedh{}+\dlreth{}+\segneth{}+CRF, nDSM corrections \\
\hline
\end{tabular}
\end{table}

\begin{table}[ht]
\centering
\caption{Per-class $F_1$-scores and overall accuracies of top
  performers on the Vaihingen benchmark (numbers copied from benchmark
  website). DLR\_7 is our ensemble model, DLR\_9 is our ensemble with
  corrected nDSM. Acronyms are taken from the official ISPRS-Vaihingen
  website.}
\label{tab:compare-participants}
\small
\begin{tabular}{|l|ccccc|c|}
\hline
& \textit{Impervious} & \textit{Building} & \textit{Low-Veg} & \textit{Tree}    & \textit{Car}     & \textit{OA} \\
\hline
DST\_2 &
90.5 \% & 93.7 \% & 83.4 \% & 89.2 \% & 72.6 \% & 89.1 \% \\
INR &
91.1 \% & 94.7 \% & 83.4 \% & 89.3 \% & 71.2 \% & 89.5 \% \\
ONE\_6 &
91.5 \% & 94.3 \% & 82.7 \% & 89.3 \% & \textbf{85.7 \%}  & 89.4 \% \\
ONE\_7 &
91.0 \% & 94.5 \% & \textbf{ 84.4 \%} & 89.9 \% & 77.8 \% & 89.8 \% \\
DLR\_7 &
91.4 \% & 93.8 \% & 83.0 \% & 89.3 \% & 82.1 \% & 89.4 \% \\
DLR\_9 &
\textbf{92.4 \%} & \textbf{95.2 \%} & 83.9 \% & \textbf{89.9 \%} & 81.2 \% & \textbf{90.3 \%} \\
\hline
\end{tabular}
\end{table}


\subsection{A word on data quality}
\label{sec:experiments-manual}

\begin{figure}[H]
\includegraphics[width=\textwidth]{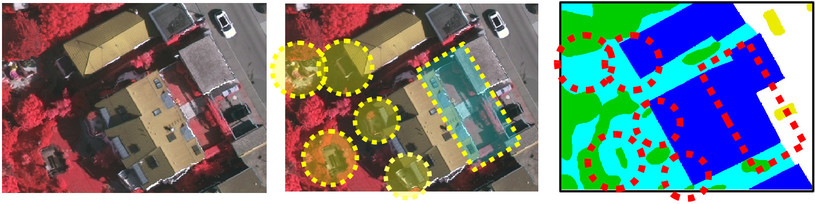}
\caption{Examples of ground truth labeling errors. Yellow/red circles
  denote missing or incorrectly delineated objects.}
\label{fig:errors-annotation}
\end{figure}

In our experiments, we repeatedly noticed inaccuracies of the ground
truth data, such as those shown in Figure~\ref{fig:errors-annotation}
(similar observations were made by
\citet{paisitkriangkrai2015effective}).
Obviously, a certain degree of uncertainty is unavoidable when
annotating data, in particular in remote sensing images with their
small objects and many boundaries.
We thus decided to re-annotate one image (\emph{image-23}) from our
Vaihingen validation set with great care,
to assess the ``inherent'' labeling
accuracy.
We did this only for the two easiest classes \emph{buildings} and
\emph{cars}, since the remaining classes have significant definition
uncertainty and we could not ensure to use exactly the same
definitions as the original annotators.

We then evaluate the new annotation, the ground truth from the
benchmark, and the output of our best model, against each other.
Results are shown in Table~\ref{tab:gt-comparison}.
One can see significant differences, especially for the \emph{cars}
which are small and have a large fraction of pixels near the
boundary. Considering the saturating progress on the benchmark
(differences between recent submissions are generally $<2$\%) there is
a very real danger that annotation errors influence the results and
conclusions.
It may be surprising, but the Vaihingen dataset (and seemingly also
the Potsdam dataset) is reaching its limits after barely 3 years of
activity. This is a very positive and tangible sign of progress, and a
strong argument in favor of public benchmarks. But it is also a
message to the community: if we want to continue using benchmarks --
which we should -- then we have to make the effort and extend/renew
them every few years.
Ideally, it may be better to move to a new dataset altogether, to
avoid overfitting and ensure state-of-the-art sensor quality.

\begin{table}[htb]
\caption{Inter-comparison between ISPRS Vaihingen ground truth, our
  own annotation, and our best models. Significant differences occur,
  suggesting that the benchmark may have to be re-annotated or
  replaced. See text for details.}
\label{tab:gt-comparison}
\centering
\begin{adjustbox}{center}
\footnotesize
\begin{tabular}{l|c|c||c|c||c|c|}
\cline{2-7}
& \multicolumn{2}{c||}{\hedh+\segneth-\textit{Msc}}
& \multicolumn{2}{c||}{\hedh+\dlreth+\segneth}
& benchmark label & our label\\
\cline{2-7}
& benchmark label & our label & benchmark label & our label & our label  & benchmark label \\
\hline
\multicolumn{1}{|c|}{\textit{Building}} &
97.3 \% & 97.8 \% & 97.7 \% & 98.0 \% & 97.9 \% & 94.7 \% \\ \hline
\multicolumn{1}{|c|}{\textit{Car}} &
84.6 \% & 88.1 \% & 79.8 \% & 83.3 \% & 93.2 \% &  88.8 \% \\ \hline
\end{tabular}
\end{adjustbox}
\end{table}

\section{Conclusion}\label{sec:conclusion}

We have developed DCNN models for semantic segmentation of
high-resolution aerial images, which explicitly represent and extract
the boundaries between regions of different semantic classes.
Empirically, including class boundaries significantly improves
different DCNN architectures, and was the single most important
performance boost in our final model, which achieves excellent
performance on the ISPRS Vaihingen and Potsdam benchmarks.

Moreover, we have presented an extensive study of semantic
segmentation architectures, including presence or absence of fully
connected layers, use of class boundaries, multi-scale processing, and
multi-network ensembles.

One aspect that we have not yet investigated, but that might be
needed to fully exploit the information in the segmentation
boundaries, are class-specific boundaries. Our current boundaries are
class-agnostic, they do not know which classes they actually separate.
It appears that this information could be preserved and used. Pushing
this idea to its extremes, it would in fact be enough to detect
\emph{only} the class boundaries, if one can ensure that they form
closed regions.

Although DCNNs are the state-of-the-art tool for semantic segmentation,
they have reached a certain degree of saturation, and further
improvements of segmentation quality will probably be small, tedious,
and increasingly problem-specific.
Nevertheless, there are several promising directions for future
research. We feel that model size is becoming an issue. Given the
excessive size and complexity of all the best-performing DCNN models,
an interesting option would be to develop methods for compressing
large, deep models into smaller, more compact ones for further
processing. First ideas in this direction have been brought up by
\citet{hinton2015distilling}.

\section*{References}\label{sec:references}


\bibliographystyle{elsarticle-harv}
\bibliography{ref}

\end{document}